\documentclass{article}

\PassOptionsToPackage{numbers,compress}{natbib}
 \usepackage[preprint]{neurips_2026}

\usepackage[utf8]{inputenc} 
\usepackage[T1]{fontenc}    
\usepackage{hyperref}       
\usepackage{url}            
\usepackage{booktabs}       
\usepackage{amsfonts}       
\usepackage{nicefrac}       
\usepackage{microtype}      
\usepackage{xcolor}         
\usepackage{listings}       
\usepackage{multirow}       
\usepackage{graphicx}       
\usepackage{makecell}       
\usepackage{subcaption}     
\usepackage{amsmath}        
\usepackage[most]{tcolorbox}
\tcbuselibrary{breakable, skins}
\newcommand{\inlinecode}[1]{{\small\texttt{#1}}}
\newcommand{\std}[2]{#1{\tiny$\pm$}{\scriptsize #2}}
\newcommand{\best}[2]{\textbf{\std{#1}{#2}}}
\newcommand{\second}[2]{\underline{\std{#1}{#2}}}

\title{MILM: Large Language Models for Multimodal Irregular Time Series with Informative Sampling}

\author{%
Hsing-Huan Chung$^{1}$ \quad
Shijun Li$^{1}$ \quad
Yoav Wald$^{2}$ \quad
Xing Han$^{3}$ \quad
Suchi Saria$^{3}$ \quad
Joydeep Ghosh$^{1}$ \\
$^{1}$University of Texas at Austin \quad
$^{2}$Technion-IIT \quad
$^{3}$Johns Hopkins University
}

\begin{document}

\maketitle

\begin{abstract}
Multimodal irregular time series (MITS) consist of asynchronous and irregularly sampled observations from heterogeneous numerical and textual channels. In healthcare, for example, patients' electronic health records (EHR) include irregular lab measurements and clinical notes. The irregular timing and channel patterns of observations carry predictive signal alongside the numerical values and textual content. LLMs are natural candidates for processing such heterogeneous data, given their extensive pretrained knowledge spanning textual and numerical domains. Existing MITS models focus on specialized fusion mechanisms without leveraging LLMs for unified MITS processing, while LLM-based approaches for irregular time series focus on purely numerical observations, leaving MITS an underexplored setting for LLMs. Moreover, no prior work has examined whether LLMs can learn to use informative sampling patterns inherent to MITS. We introduce \textbf{MILM} (\textbf{M}ultimodal \textbf{I}rregular time series \textbf{L}anguage \textbf{M}odel), which represents MITS as time-ordered triplets in Extensible Markup Language (XML) format and fine-tunes an LLM through a two-stage strategy for MITS classification. The first stage trains on value-redacted MITS to predict from sampling patterns alone, and the second stage trains on full MITS to jointly model sampling patterns and observed values. Our two-stage model (MILM-2S) and its single-stage counterpart (MILM-Direct) achieve the best and second-best average performance on multiple EHR datasets. Further value redaction evaluations confirm that sampling patterns carry predictive signal and that MILM-2S learns to exploit them. In the value pending evaluation we introduce, where some values are unavailable at prediction time, MILM-2S outperforms MILM-Direct by a larger margin compared to standard evaluation. For MILM-2S, preserving the time and channel of value-pending observations as additional sampling information further improves in-hospital mortality prediction. 
\end{abstract}


\section{Introduction}

Multimodal irregular time series (MITS)~\cite{chang2025timeimm} consist of asynchronous and irregularly sampled observations from heterogeneous channels spanning numerical and textual modalities. MITS are prevalent in healthcare~\cite{zhang2023improving}, finance~\cite{koval2025multimodal}, and system monitoring~\cite{zhang2024survey} among others. For example, in electronic health records (EHR) \cite{johnson2023mimic,PhysioNet-mimiciv-3.1,pollard2018eicu,PhysioNet-eicu-crd-2.0}, a patient's stay comprises lab measurements and clinical notes irregularly ordered and authored over time due to their on-demand nature. Unlike regularly sampled time series, which arise from periodic measurement of data processes, MITS are jointly governed by data processes and sampling processes, where the latter determine when and which channels are observed. This dual-process nature introduces irregularity in the timing and channel patterns of observations, making them informative signals beyond the observed values. Combined with the heterogeneity of modalities, this makes MITS a richer yet more challenging source of predictive information.

In this work, we focus on MITS classification motivated by clinical prediction tasks~\cite{zhang2023improving,khadanga2019using,han2024fusemoe}. A canonical example is in-hospital mortality prediction, where a model processes a patient's lab measurements and clinical notes to predict whether the patient will survive their hospital admission. Accurate prediction requires interpreting numerical lab values alongside clinical narratives, while also exploiting when and which lab tests are ordered and clinical notes are authored as informative signals about the patient's evolving condition. LLMs are natural candidates for processing such heterogeneous observations, given their extensive pretrained knowledge spanning both textual and numerical domains. Existing MITS models~\cite{zhang2023improving,khadanga2019using,han2024fusemoe} focus on specialized multimodal fusion mechanisms, leaving the use of LLMs for unified processing of the temporal structure, numerical values, and textual content underexplored. Meanwhile, LLM-based approaches for irregular time series~\cite{kwon2025mind,zhang2025unleashing} focus on purely numerical observations without the textual modalities that make MITS a particularly natural fit for LLMs. Moreover, no prior work has examined whether and how LLMs can leverage sampling patterns in MITS as predictive signal in their own right. To this end, we develop an LLM-based framework for classifying MITS with informative sampling.

We introduce \textbf{MILM} (\textbf{M}ultimodal \textbf{I}rregular time series \textbf{L}anguage \textbf{M}odel), a framework that serializes MITS into time-ordered triplets in Extensible Markup Language (XML) format and adapts an LLM via a two-stage fine-tuning strategy. Time-ordered serialization preserves the causal structure of the data and sampling processes, while the XML format makes the structured nature of MITS explicit. MILM first performs value redaction fine-tuning, teaching the model to predict from sampling patterns alone, and then continues fine-tuning on the full data to jointly utilize observed values and the learned sampling patterns. Empirically, the two-stage variant, \textbf{MILM-2S}, achieves the best average rank across all datasets and metrics, while the variant directly fine-tuned on the full data, \textbf{MILM-Direct}, ranks second overall. We further evaluate the models under value redaction and show that sampling patterns carry predictive signal, that Stage 1 effectively captures this signal, and that Stage 2 retains most of this capability while still outperforming both the off-the-shelf LLM and MILM-Direct. We also introduce a realistic value pending evaluation, where some values are unavailable at prediction time. MILM-2S consistently outperforms MILM-Direct under both countermeasures of dropping value-pending observations and revealing their time and channel. For MILM-2S, preserving the time and channel of value-pending observations as additional sampling information further improves in-hospital mortality prediction on MIMIC-IV. In summary, our contributions are:
\begin{itemize}
    \item We present the first study of LLMs for MITS classification and the first investigation of whether and how LLMs can learn to exploit informative sampling patterns in this setting.
    \item We propose \textbf{MILM}  (\textbf{M}ultimodal \textbf{I}rregular time series \textbf{L}anguage \textbf{M}odel), which represents MITS as time-ordered triplets in XML format and applies two-stage fine-tuning to first learn from sampling patterns and then jointly model sampling patterns and values.
    \item We present extensive experiments on EHR-derived MITS datasets, showing that MILM achieves strong performance, while additional value redaction and value pending evaluations demonstrate that it can effectively learn and use informative sampling patterns.
\end{itemize}



\section{Related Work}

\subsection{Irregular Time Series}

Irregular time series \citep{shukla2020survey} arise when observations arrive at non-uniform time intervals and different variables are recorded asynchronously. One common approach is to represent each sample with aligned value, mask, and time-gap matrices, allowing sequence models to exploit observed values and missingness patterns \cite{che2018recurrent}. Other methods such as IP-Net \cite{shukla2018interpolationprediction}, mTAND \cite{shukla2021multitime}, and ATENet \cite{leeadaptive} learn interpolation representations at reference times. Another line of work explicitly models continuous-time latent dynamics with ODEs \cite{rubanova2019latent,de2019gru}, CDEs \cite{kidger2020neural}, and SDEs \cite{schirmer2022modeling,ansari2023neural}. Several recent approaches adapt attention \cite{chen2023contiformer,zhang2023warpformer} or recast irregular time series as sets \cite{horn2020set}, graphs \cite{zhang2022graphguided,zhang2024irregular,yalavarthi2024grafiti}, or images \cite{li2023time} to address asynchronous measurements and cross-variable dependencies. The above approaches do not consider the multimodal setting where channels may also include free text.



\subsection{Multimodal Irregular Time Series}

Multimodal irregular time series (MITS) \cite{chang2025timeimm} extend the irregular time series setting beyond purely numerical observations to include additional modalities such as text. \citet{khadanga2019using} use timestamped clinical notes with numerical time series for ICU prediction, but handle the irregular note stream with simple temporal weighting. UTDE \cite{zhang2023improving} uses mTAND \cite{shukla2021multitime} to model both irregular numerical observations and timestamped clinical note embeddings. The resulting representations of the text and numerical channels are then fused with self- and cross-attention blocks for prediction. FuseMoE \cite{han2024fusemoe} further extends multimodal fusion to fleximodal settings with additional modalities and mixture-of-experts routing. Time-IMM \cite{chang2025timeimm} is a recent benchmark for forecasting on irregular multimodal time series. In contrast, we study time-series-wise MITS classification where each full trajectory is itself a labeled sample, in line with prior clinically motivated multimodal prediction settings \cite{zhang2023improving,khadanga2019using,han2024fusemoe}. To the best of our knowledge, prior work has not studied the use of LLMs for jointly processing the temporal structure, numerical values, and textual content under this setting.

\subsection{LLM for Time Series}
Prior LLM-for-time-series methods have largely focused on regularly sampled numerical series, either by serializing values into text for direct prompting \cite{xue2023promptcast,gruver2023large} or by aligning numerical sequences to the LLM representation space \cite{cao2023tempo,zhou2023one,jin2024timellm,sun2023test,liu2024autotimes,pan2024s,liu2025calf,liu2025timecma}. There is also a line of work on multimodal settings \cite{belyaeva2023multimodal,chan2024medtsllm,jia2024gpt4mts,cheng2025instructime,wang2025chattime,lee2025timecap,langer2025opentslm} where time series are combined with text used as context or summaries rather than as observations within the temporal sequence. Adjacent work on time-series foundation models \cite{das2023decoder,shi2024time,ansari2024chronos,ansari2025chronos} also relies on large transformer architectures, but is pretrained on time-series corpora rather than natural language.

Aside from regularly sampled time series, LLMs have also been applied to continuous-time event streams \cite{liu2024tpp,gupta2025last,kong2026byte}. These works focus on temporal point process problems such as next-event time/type prediction and intensity estimation, whereas we study time-series-wise prediction from timestamped channel-value observations whose values can be numerical or textual. On the other hand, LLMs for purely numerical irregular time series \cite{kwon2025mind,zhang2025unleashing} have also been explored. VITAL \cite{kwon2025mind} extends Time-LLM \cite{jin2024timellm}, a representation alignment approach for regularly sampled time series, to handle missingness. ISTS-PLM \cite{zhang2025unleashing} uses time, variable, and value embedders to encode irregular numerical observations before processing them with a pretrained language model. In contrast, we study LLMs for multimodal irregular time series, where observations arrive asynchronously across channels and their values may be either numerical measurements or unstructured text.


\subsection{Informative Presence/Missingness/Sampling}

Informative presence or missingness \cite{sisk2021informative,tan2023informative} refers to the setting where the presence or absence of data is itself informative because data are often missing not at random \cite{little2019statistical}. Informative sampling \cite{lin2004analysis,vanderschueren2023accounting} extends this idea to time: the fact that measurements are taken at particular times can also carry signal, e.g., increasingly frequent laboratory tests may indicate worsening health status. Prior work has examined informative sampling as a source of bias in longitudinal regression \cite{lin2004analysis,gasparini2020mixed}, treatment effect forecasting \cite{vanderschueren2023accounting}, and time-to-event modeling under changing sampling policies \cite{jeanselme2025prediction}. Related ideas also appear in irregular time series models that treat sampling patterns as part of the predictive signal, for example through measurement intensity \cite{shukla2018interpolationprediction,rubanova2019latent} or elapsed time and missingness indicators \cite{che2018recurrent}. More recently, \citet{kobayashi2025mind} study the impact of informative missingness on LLM predictions from static data. In contrast, we study whether LLMs can learn and use informative sampling patterns in the temporal setting of multimodal irregular time series.

\section{Problem Formulation}

\subsection{Multimodal Irregular Time Series}
We consider multimodal irregular time series (MITS) with observations at arbitrary time points within a time range $\mathcal{I}_T = [T_{\text{start}}, T_{\text{end}}]$. We denote the set of channel types by $\mathcal{C}$. In the EHR example (Fig.~\ref{fig:irregular_patient_trajectory}), $\mathcal{C}$ could contain numerical laboratory test channels and textual clinical note channels. For each channel $c \in \mathcal{C}$, we consider two interacting processes.

\textbf{Data Process.} Let $X_c(t): \mathcal{I}_T \to \mathcal{X}_c$ denote the value of channel $c$ at time $t$, where $\mathcal{X}_c$ is the channel value space (e.g., $\mathbb{R}$ for numerical measurements, a string space for clinical notes). This process is only observed when channel $c$ is sampled.

\textbf{Sampling Process.} Let $N_c(t): \mathcal{I}_T \to \mathbb{N}$ be a counting process recording the cumulative number of observations of channel $c$ up to time $t$. It increments by 1 whenever channel $c$ is observed: $dN_c(t) = 1$ if an observation occurs at time $t$, and $dN_c(t)=0$ otherwise, where $dN_c(t) = N_c(t) - \lim_{s \to t^-} N_c(s)$. The sampling processes may depend on the data processes.

For each channel $c$, let $t_1^{(c)} < \cdots < t_{n_c}^{(c)}$ denote the times at which $dN_c(t) = 1$, i.e. the $n_c = N_c(T_\text{end})$ observation times of channel $c$. The per-channel observation sequence is $\mathbf{s}_c = ((t_k^{(c)}, x_k^{(c)}))_{k=1}^{n_c}$, where $x_k^{(c)}= X_c(t_k^{(c)})$ is the value or content. The full MITS is the collection $\{\mathbf{s}_c | c \in \mathcal{C}\}$.


\subsection{Multimodal Irregular Time Series Classification}

\begin{figure}[t]
    \centering

    \begin{subfigure}[t]{0.54\linewidth}
        \centering
        \includegraphics[width=\linewidth]{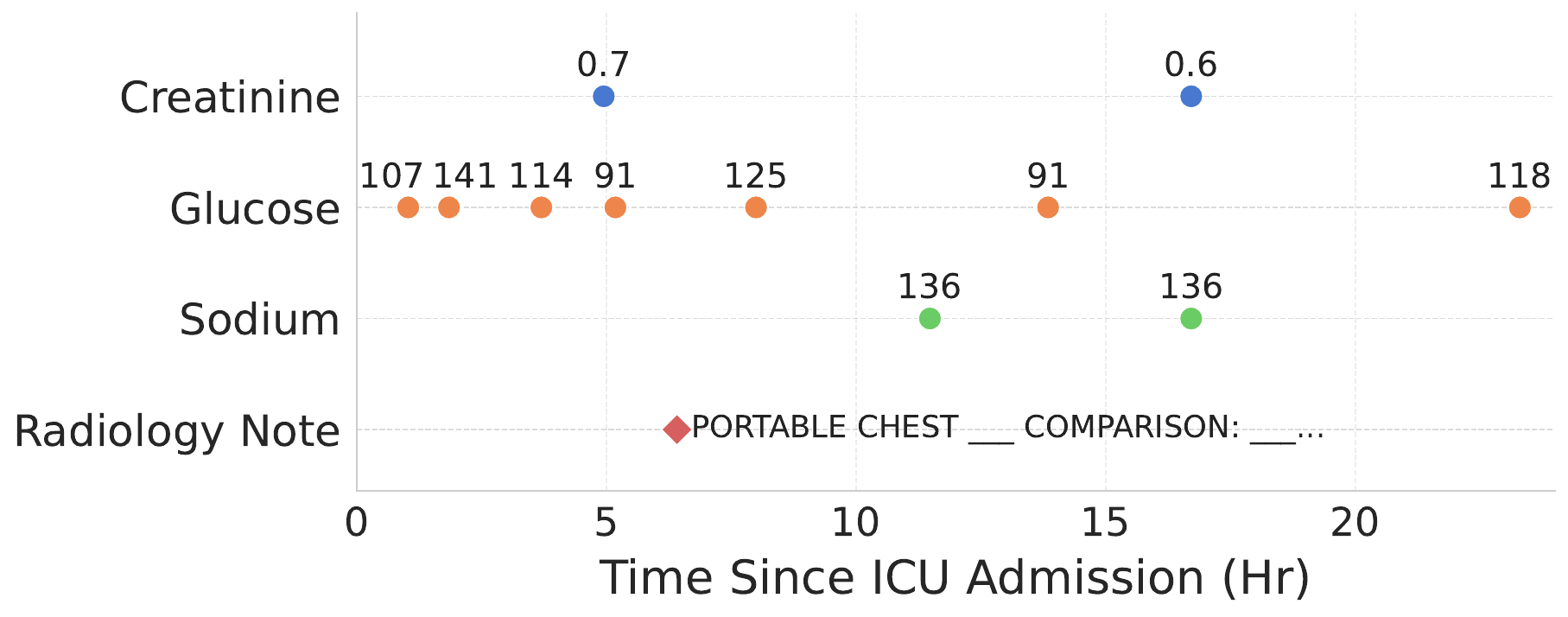}
        \caption{}
        \label{fig:irregular_patient_trajectory}
    \end{subfigure}
    \hfill
    \begin{subfigure}[t]{0.22\linewidth}
        \centering
        \includegraphics[width=\linewidth]{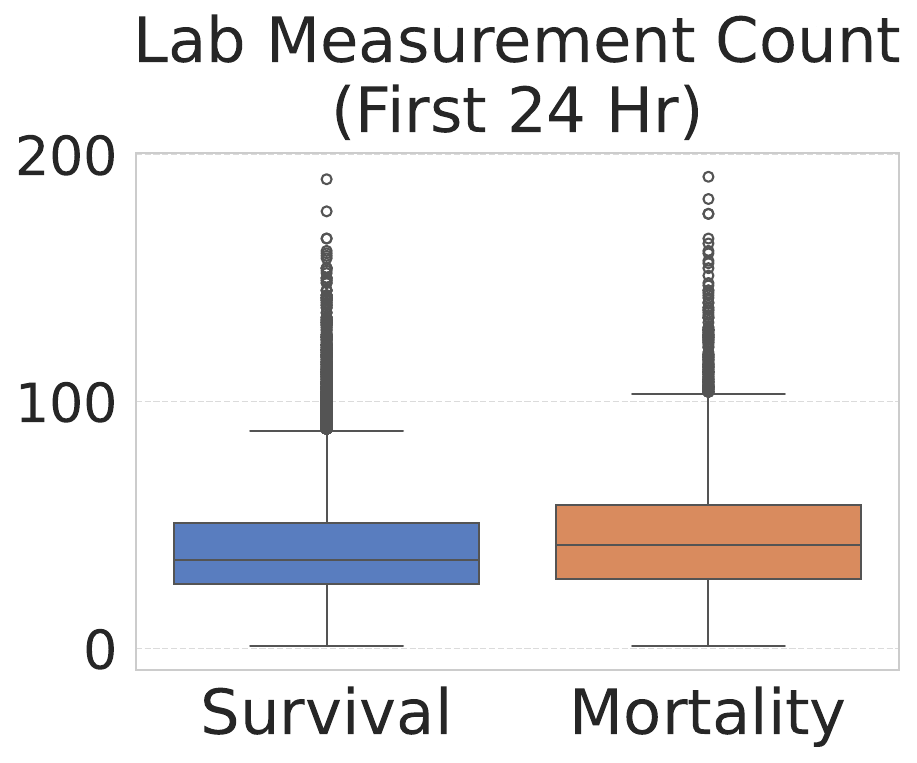}
        \caption{}
        \label{fig:lab_measurement_stats}
    \end{subfigure}
    \hfill
    \begin{subfigure}[t]{0.22\linewidth}
        \centering
        \includegraphics[width=\linewidth]{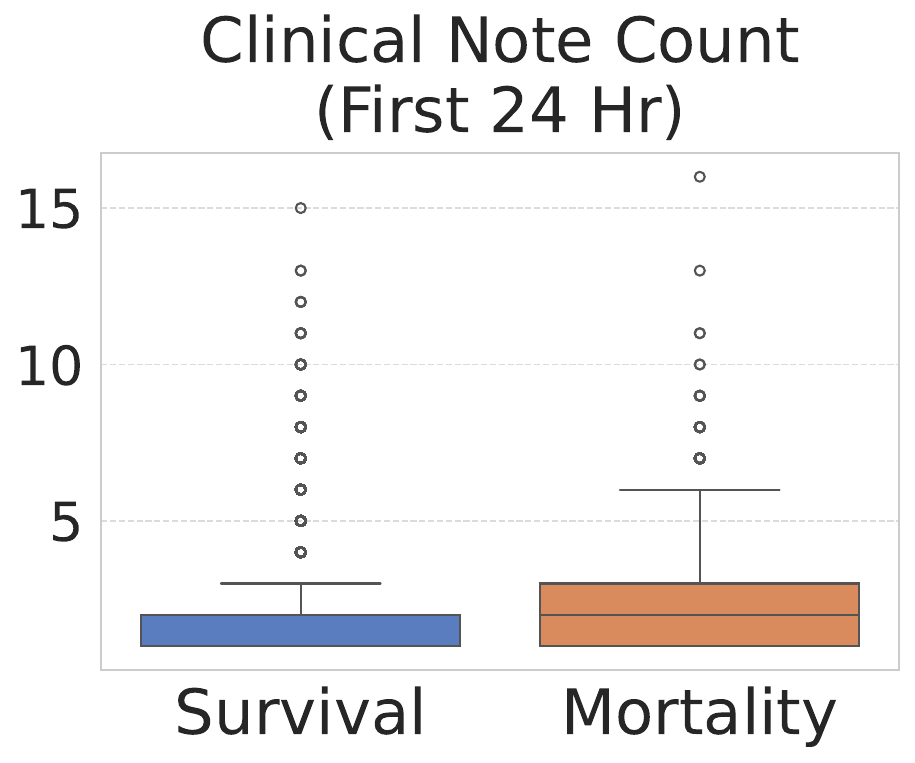}
        \caption{}
        \label{fig:note_measurement_stats}
    \end{subfigure}

    \caption{
        ~\ref{fig:irregular_patient_trajectory}: A patient's lab measurement and clinical note trajectory from the first 24 hours of an ICU stay, drawn from the MIMIC-IV-Demo database~\cite{PhysioNet-mimic-iv-demo-2.2}.
        ~\ref{fig:lab_measurement_stats} and ~\ref{fig:note_measurement_stats}: Distributions of numbers of lab measurements and clinical notes within $24$ hours grouped by outcome. Although counts are rough summaries of sampling patterns, mortality cases already exhibit more frequent observations. The full sampling patterns carry finer-grained signals that may be even more predictive of the outcome.
    }
    \label{fig:irregular_sampling_example}
\end{figure}
        

Let $Y \in \{1, \ldots, K\}$ denote the label of a MITS sample. We observe a dataset $\mathcal{D} = \{\mathcal{S}_i\}_{i=1}^M$ consisting of $M$ i.i.d. realizations of the random observed data object $\mathcal{S} = (\{\mathbf{s}_c | c \in \mathcal{C}\}, Y)$. The goal is to learn a MITS classifier $f$ that minimizes the expected classification loss $\mathbb{E}_{\mathcal{S}}[l(f(\{\mathbf{s}_c | c \in \mathcal{C}\}), Y)]$, where $l$ could be the 0/1 loss or cross-entropy loss. Each $\mathbf{s}_c$ encodes two sources of information: (i) the observed values ${x_k^{(c)}}$ that directly reflect the data process and (ii) the sampling patterns ${t_k^{(c)}}$ and $n_c$, which are also informative about $Y$ because sampling may depend on both the observed and unobserved parts of the data process. For example, a patient with a critically abnormal lab result or a worsening clinical state not yet documented in the system may be assigned more follow-up tests. Indeed, in the MIMIC-IV EHR database ~\cite{johnson2023mimic,PhysioNet-mimiciv-3.1,PhysioNet-mimic-iv-note-2.2}, patient stays resulting in in-hospital mortality exhibit more lab measurements and clinical notes than surviving stays (Fig.~\ref{fig:lab_measurement_stats} and ~\ref{fig:note_measurement_stats}), though the total count is merely a coarse summary. The full sampling pattern that encompasses timings, channel identities, and orderings encodes even richer information. An effective MITS classifier should make use of both observed values and sampling patterns for prediction.

\section{Method}\label{sec:method}

We introduce \textbf{MILM}: \textbf{M}ultimodal \textbf{I}rregular time series \textbf{L}anguage \textbf{M}odel (Fig.~\ref{fig:milm}), a framework that serializes MITS into time-ordered triplets in XML format and employs a two-stage fine-tuning strategy to adapt an LLM to exploit both sampling patterns and values for MITS classification.

\subsection{MITS Representation}\label{subsec:mits_repr}
\begin{figure}[t]
\centering
\includegraphics[width=0.95\columnwidth]{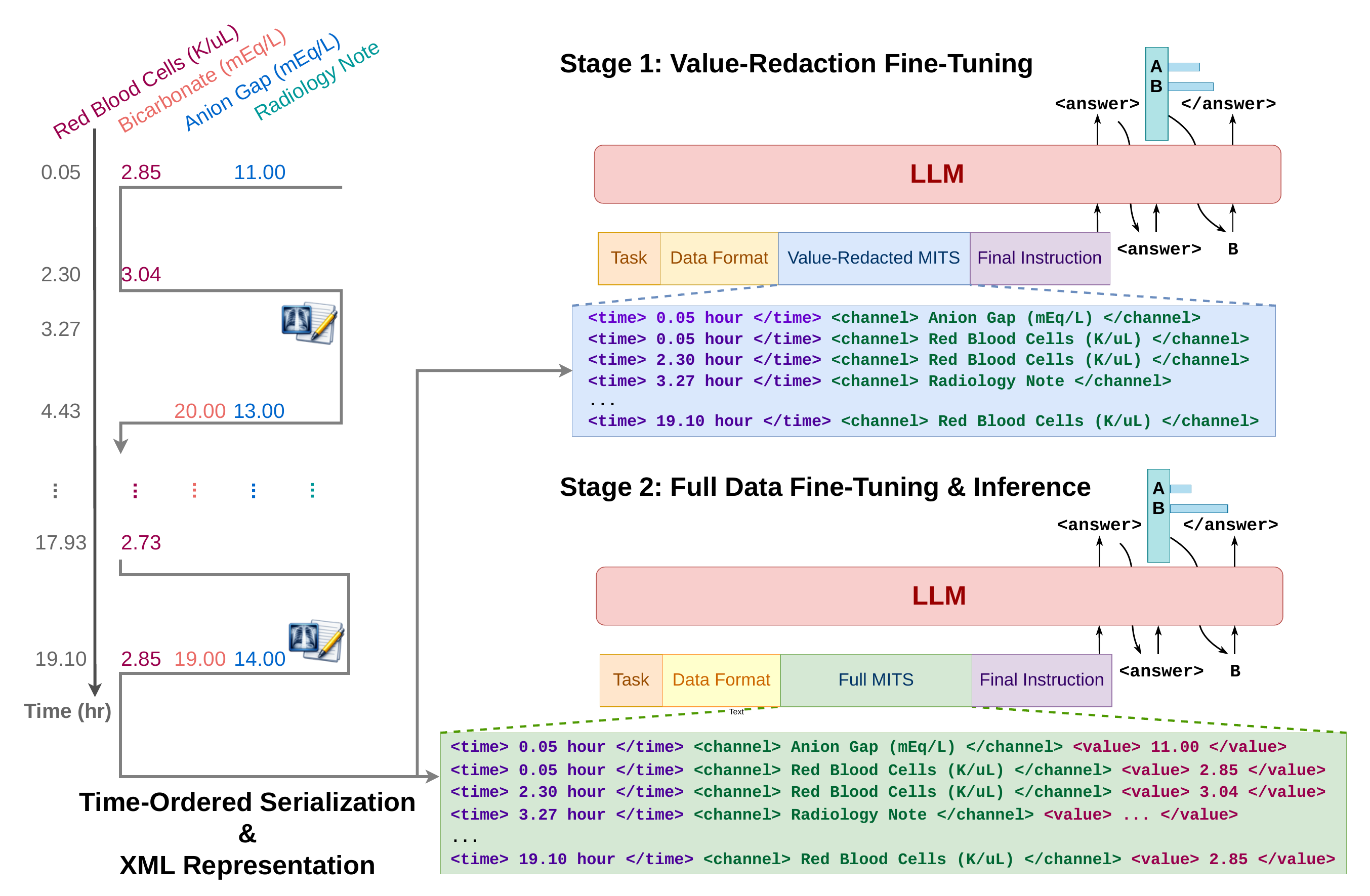}
\caption{An illustration of MILM. MILM serializes the MITS to a time-ordered XML representation and fine-tunes the LLM in two stages. Value-redaction fine-tuning first teaches the model to predict from sampling patterns alone, while full-data fine-tuning then incorporates value information on top of the developed sampling-pattern knowledge. During inference, MILM uses the Stage 2 prompt format and extracts the answer-token logits as prediction scores.}
\label{fig:milm}
\end{figure}

We design a MITS representation that enables LLMs to process numerical and textual observations.
Prior work on LLMs for irregular time series \cite{kwon2025mind,zhang2025unleashing} views time series observations as numerical objects and transforms them into a language model's token embedding space, using representations foreign to the LLM's pretraining. This limits the model's ability to draw on pretrained knowledge. We instead represent MITS directly in the LLM's native token vocabulary, enabling joint processing of numerical and text modalities without representational alignment. We explain the process below:

\paragraph{Time-Ordered Serialization.}
As shown in Fig.~\ref{fig:milm}, a MITS $\{\mathbf{s}_c | c \in \mathcal{C}\}$ is a sparse 2-D object indexed by channel and time. We apply time-ordered serialization, augmenting each observation $(t_k^{(c)}, x_k^{(c)}) \in \mathbf{s}_c$ with its channel label $c$ and globally sorting by time, breaking ties by predefined channel order, to obtain a flattened sequence $((t_i, c_i, x_i))_{i=1}^n$, where $t_i$, $c_i$, and $x_i$ are the time, channel, and value of the $i$'th observation, $n = \sum_{c \in \mathcal{C}} n_c$, and $t_i \leq t_j$ for all $i < j$. Preserving global temporal order ensures that the causal structure of the sampling and data processes is directly reflected in the sequence and accessible to the model.

\paragraph{XML-Formatted Triplet Representation.}
Each observation $(t_i, c_i, x_i)$ in the serialized sequence has three semantically distinct components: when it occurred, which channel was observed, and what value was recorded. We represent each observation as an XML-formatted triplet with tags \inlinecode{<time>}, \inlinecode{<channel>}, and \inlinecode{<value>}, explicitly encoding this three-part structure so that the boundary between observations and the role of each component remains unambiguous to the model. Consecutive observations are separated by a newline. For example, the following snippet specifies a MITS with $3$ observations from the glucose, anion gap, and radiology note channels:

\begin{lstlisting}[basicstyle=\scriptsize\ttfamily, columns=flexible, breaklines=false, frame=none,
backgroundcolor=\color{gray!10},
xleftmargin=12pt, xrightmargin=4pt,
aboveskip=3pt, belowskip=3pt]
<time> 0.88 hours </time> <channel> Glucose (mg/dL) </channel> <value> 170.00 </value>
<time> 3.80 hours </time> <channel> Anion Gap (mEq/L) </channel> <value> 11.00 </value>
<time> 4.00 hours </time> <channel> Radiology Note </channel> <value> Portable chest... </value>
\end{lstlisting}

This representation has two advantages. First, expressing observations directly in the LLM's native vocabulary allows pretrained token embeddings to apply without representation alignment. Preserving timings and channel identities as natural language additionally grounds each value in semantic context, enabling the model to draw on pretrained knowledge when interpreting it. Second, the XML format signals that the input is a data object indexed by time and channel rather than free-form text, preserving the structured nature of MITS within a text-based representation.


\subsection{LLM For MITS Classification}\label{subsec:llm4mits}
We describe how we use LLMs for MITS classification, covering prompt and target construction for both inference and supervised fine-tuning (SFT), and prediction score extraction at inference time.

\paragraph{Prompt and Target Construction.} To condition the LLM for the prediction task, the prompt consists of three specification blocks followed by the serialized MITS and a final instruction. The \textbf{task specification} describes the data domain and the prediction target, e.g., ``timestamped observation from an ICU stay'' and ``predict whether the patient will survive their hospital admission''. The \textbf{data format specification} explains the triplet XML format and the semantics of each tag (e.g., the \inlinecode{<time>} tag records hours since the start of the ICU stay). The \textbf{output format specification} maps each class label to an alphabet letter and instructs the model to place its prediction in an \inlinecode{<answer>} tag (e.g., \inlinecode{<answer> B </answer>}). See Appendix~\ref{app:prompt} for the full prompt templates. For SFT, the training target is the ground-truth class letter and the enclosing \inlinecode{<answer>} tags. The model is trained via next-token prediction loss on the full target sequence.

\paragraph{Prediction Score Extraction.} Following \citet{robinson2023leveraging}, at inference we extract prediction scores from the logits at the position immediately following the \inlinecode{<answer>} open tag and apply softmax over the class letter tokens to obtain a probability distribution. We prefer logit-based extraction over verbalizing class probabilities because logits faithfully reflect the model's class assignment belief, whereas verbalizing interposes an extra decoding step. Also, verbalizing probability scores requires numerical confidence targets during SFT, which are typically unavailable.

\subsection{Two-Stage Fine-Tuning}

An effective MITS classifier should exploit signals from both values and sampling patterns. However, the signal in sampling patterns is subtler than that in values, particularly when text channels are present and value tokens constitute a large portion of the sequence. To avoid bypassing sampling patterns in favor of the more immediately apparent value signal, we propose a two-stage fine-tuning strategy that forces sampling-pattern representations to develop before value information is introduced.

\paragraph{Stage 1: Value Redaction Fine-Tuning.} In the first stage, we fine-tune the LLM on MITS with all values redacted. Each observation is presented as a time-channel pair, as in the following example:
\begin{lstlisting}[basicstyle=\scriptsize\ttfamily, columns=flexible, breaklines=false, frame=none,
backgroundcolor=\color{gray!10},
xleftmargin=12pt, xrightmargin=4pt,
aboveskip=3pt, belowskip=3pt]
<time> 0.88 hours </time> <channel> Glucose (mg/dL) </channel>
<time> 3.80 hours </time> <channel> Anion Gap (mEq/L) </channel>
<time> 4.00 hours </time> <channel> Radiology Note </channel>
\end{lstlisting}
The data format specification in the prompt is adjusted to reflect the redacted format. An additional note clarifies that values were recorded but withheld and instructs the model to rely on timing and channel patterns for prediction. Stage 1 trains the model to develop predictive representations of the sampling processes $\{N_c(t) | c \in \mathcal{C}\}$ through the timings and channel identities of observations.

\paragraph{Stage 2: Full Data Fine-Tuning.} In the second stage, we continue fine-tuning from the Stage 1 checkpoint on the full MITS data, using the same representation and prompt construction described in Sections~\ref{subsec:mits_repr} and~\ref{subsec:llm4mits}. Building on the sampling-pattern representations developed in Stage 1, Stage 2 trains the model to jointly exploit both the data processes $\{X_c(t) | c \in \mathcal{C} \}$ and the sampling processes $\{N_c(t) | c \in \mathcal{C}\}$. 
The Stage 2 model is used for the final prediction.

This two-stage scheme is enabled by the text-based representation: removing the \inlinecode{<value>} tokens preserves the presence of each observation while unambiguously signaling that its value is withheld, allowing the same model architecture to process both redacted and full MITS without modification.


\section{Experiments}

\subsection{Setup}\label{subsec:setup}
\paragraph{Data.} We conduct experiments on datasets extracted from two major public EHR databases: MIMIC-IV \cite{johnson2023mimic,PhysioNet-mimiciv-3.1,PhysioNet-mimic-iv-note-2.2} and eICU Collaborative Research Database \cite{pollard2018eicu,PhysioNet-eicu-crd-2.0}. Similar to \citet{jeanselme2025prediction}, we extract the laboratory tests as the numerical channels due to their inherent irregular sampling nature. In addition, we extract the clinical notes as the text channel. On each database, we develop two datasets corresponding to the in-hospital mortality (IHM) and length-of-stay (LOS) prediction tasks \cite{harutyunyan2019multitask,sheikhalishahi2020benchmarking} given the first $24$ hours of laboratory tests and clinical notes of the ICU stay. We describe the datasets in detail in Appendix~\ref{app:dataset}.

\paragraph{Baselines.}
We select $3$ classes of irregular time series model baselines: (i) Non-LLM-based: GRU-D \cite{che2018recurrent}, mTAND \cite{shukla2021multitime}, and VITAL-stats \cite{kwon2025mind}  (ii) LLM-based: VITAL-LLM \cite{kwon2025mind} and ISTS-PLM \cite{zhang2025unleashing}, and (iii) Multimodal: UTDE \cite{zhang2023improving} and FuseMoE \cite{han2024fusemoe}. Since the datasets belong to the biomedical domain, we use BioBERT \cite{lee2020biobert} as the note embedding model. The unimodal baselines do not use the text channel by default. For those models, we implement a ``+ Note'' version that uses the timestamp-to-text fusion (TTF) module proposed by \citet{chang2025timeimm} to learn the text channel representation to be concatenated with the numerical channel representation before the prediction head.

\paragraph{MILM Setup.}
We fine-tune Qwen3-4B-Instruct-2507 \cite{yang2025qwen3} and experiment with two variants of MILM: \textbf{MILM-Direct}, which directly does full data fine-tuning, and \textbf{MILM-2S}, which applies the two-stage fine-tuning strategy described in Section~\ref{sec:method}. Both variants use QLoRA \cite{dettmers2023qlora} with rank $r=16$, $\alpha=16$, and dropout $0.05$ applied to the query, key, value, and output projection matrices of the LLM. We use a learning rate of $10^{-4}$ with a cosine schedule and warmup ratio $0.05$. Stage 1 of MILM-2S trains for $8$ epochs. Stage 2 initializes from the best Stage 1 checkpoint, and the number of training epochs is dataset-dependent. At inference, we use SGLang \cite{zheng2024sglang} to serve the fine-tuned model in bfloat16 precision with logit-based prediction score extraction. We also evaluate the off-the-shelf Qwen3-4B-Instruct-2507 (hereafter Qwen3-4B) as a zero-shot baseline. See Appendix~\ref{app:model_config} for further details regarding implementation and hyper-parameters.

\paragraph{Evaluation.}
Each dataset is split into $70\%/15\%/15\%$ for train/validation/test. We use Area Under the Receiver Operating Characteristic Curve (AU-ROC) and Average Precision (AP) to evaluate the quality of the classification scores and report the average and standard deviation over $5$ runs.

\begin{table}[t]
\centering
\small
\setlength{\tabcolsep}{4pt}
\caption{Results on the in-hospital mortality (IHM) and length-of-stay (LOS) prediction datasets extracted from MIMIC-IV and eICU. Best results are \textbf{boldfaced} and second-best results are \underline{underlined}.}
\label{tab:main_results}
\scalebox{0.9}{
\begin{tabular}{lccccccccc}
\toprule
& \multicolumn{2}{c}{\textbf{MIMIC-IV-IHM}}
& \multicolumn{2}{c}{\textbf{MIMIC-IV-LOS}}
& \multicolumn{2}{c}{\textbf{eICU-IHM}}
& \multicolumn{2}{c}{\textbf{eICU-LOS}}
& \multirow{2}{*}{\begin{tabular}{@{}c@{}}\textbf{Avg.}\\\textbf{Rank}\end{tabular}} \\
\cmidrule(lr){2-3}
\cmidrule(lr){4-5}
\cmidrule(lr){6-7}
\cmidrule(lr){8-9}
\textbf{Method} & AU-ROC & AP & AU-ROC & AP & AU-ROC & AP & AU-ROC & AP & \\
\midrule
GRU-D                & \std{75.91}{1.03} & \std{54.08}{1.02} & \std{70.08}{0.83} & \std{46.89}{1.17} & \std{66.04}{0.99} & \std{34.98}{3.05} & \std{60.48}{2.14} & \std{51.13}{3.09} & 10.5 \\
\quad {\footnotesize + Note}        & \std{74.52}{1.71} & \std{52.28}{4.13} & \std{70.93}{2.54} & \std{47.84}{2.45} & \std{66.85}{1.21} & \std{35.97}{1.18} & \std{57.40}{3.51} & \std{45.84}{4.02} & 12.0 \\
mTAND               & \std{77.38}{0.90} & \std{56.52}{1.55} & \std{70.98}{0.96} & \std{48.99}{1.68} & \std{67.13}{1.35} & \std{36.42}{3.11} & \std{58.10}{1.77} & \std{50.36}{1.29} & 8.9 \\
\quad {\footnotesize + Note}        & \std{77.90}{1.16} & \std{58.57}{0.68} & \std{74.30}{0.82} & \std{54.26}{1.76} & \std{66.99}{1.33} & \std{37.11}{2.52} & \std{64.79}{1.03} & \std{52.20}{0.64} & 5.9 \\
UTDE                & \std{75.23}{3.10} & \std{56.14}{3.72} & \std{75.68}{2.78} & \std{56.50}{3.53} & \best{72.98}{0.92} & \std{41.58}{1.16} & \std{63.96}{3.05} & \std{50.35}{4.23} & 5.6 \\
FuseMoE             & \std{76.21}{1.45} & \std{56.27}{0.78} & \std{76.13}{0.24} & \std{57.03}{1.62} & \std{70.57}{0.75} & \std{40.43}{1.27} & \second{65.37}{2.15} & \std{52.48}{3.29} & 4.6 \\
VITAL-stats         & \std{76.99}{0.66} & \std{54.06}{2.06} & \std{73.62}{0.96} & \std{52.02}{2.89} & \std{70.51}{1.48} & \std{39.36}{2.68} & \std{60.01}{2.42} & \std{48.12}{1.56} & 8.8 \\
\quad {\footnotesize + Note}       & \std{77.80}{0.68} & \std{54.32}{1.67} & \std{73.99}{0.98} & \std{53.83}{3.61} & \std{69.77}{1.05} & \std{38.54}{2.01} & \std{59.83}{1.07} & \std{48.67}{1.42} & 8.0 \\
VITAL-LLM           & \std{66.43}{2.23} & \std{39.61}{2.45} & \std{67.60}{1.29} & \std{45.94}{2.75} & \std{61.92}{2.72} & \std{28.68}{2.44} & \std{60.14}{2.04} & \std{50.14}{2.51} & 13.1 \\
\quad {\footnotesize + Note}        & \std{67.02}{2.34} & \std{39.22}{3.20} & \std{67.97}{0.99} & \std{47.49}{1.52} & \std{61.19}{2.51} & \std{28.09}{2.53} & \std{59.91}{2.13} & \std{49.63}{1.64} & 13.4 \\
ISTS-PLM            & \std{80.49}{0.79} & \std{60.10}{1.14} & \std{73.92}{0.87} & \std{54.04}{1.77} & \std{68.93}{0.72} & \std{36.32}{1.18} & \std{62.66}{1.33} & \std{52.79}{1.44} & 5.8 \\
\quad {\footnotesize + Note}        & \std{80.07}{0.52} & \std{59.06}{0.62} & \std{73.68}{0.61} & \std{52.09}{1.13} & \std{69.02}{0.65} & \std{36.12}{0.32} & \std{62.51}{1.39} & \std{52.82}{1.38} & 6.4 \\
Qwen3-4B
                    & \std{69.82}{0.00} & \std{48.36}{0.00} & \std{67.98}{0.00} & \std{44.82}{0.00} & \std{64.53}{0.00} & \std{29.43}{0.00} & \std{55.71}{0.00} & \std{45.58}{0.00} & 13.8 \\
\midrule
MILM-Direct                & \second{80.82}{0.63} & \second{63.69}{1.29} & \second{78.24}{0.60} & \second{57.43}{2.01} & \std{71.08}{0.69} & \second{41.77}{1.72} & \std{64.97}{1.42} & \second{55.76}{1.08} & \underline{2.3} \\
MILM-2S      & \best{81.30}{0.70} & \best{63.86}{1.47} & \best{78.82}{0.17} & \best{57.45}{1.14} & \second{72.47}{1.49} & \best{42.27}{2.10} & \best{66.05}{1.79} & \best{56.22}{1.20} & \textbf{1.1} \\
\bottomrule
\end{tabular}
}
\end{table}
\subsection{Main Results}

Table~\ref{tab:main_results} summarizes performance across all four datasets. MILM-2S achieves the best average rank of $1.1$ across all eight metric-dataset combinations, with MILM-Direct a close second at $2.3$. Among the baselines, the native multimodal models FuseMoE (avg.\ rank $4.6$) and UTDE (avg.\ rank $5.6$) outperform unimodal ones. Their advantage highlights the importance of well-designed multimodal fusion strategies. In contrast, simply augmenting unimodal models with a separately computed text representation via concatenation (``+ Note'') often fails to help and sometimes hurts performance, suggesting that the integration strategy matters as much as having the text channel at all. MILM takes a complementary approach. By interleaving numerical and text observations in a time-ordered serialization with causal structure preserved, the LLM jointly attends to both modalities, achieving the best overall performance.

Comparing the two MILM variants against the zero-shot Qwen3-4B baseline illuminates the contribution of each training stage. Qwen3-4B achieves above-random performance (e.g., AU-ROC of $69.82$ on MIMIC-IV-IHM), confirming that our constructed prompt conditions the LLM to effectively leverage its pretrained knowledge for the task even without fine-tuning. Yet it ranks last overall (avg.\ rank $13.8$), indicating that domain-specific MITS patterns require further adaptation. MILM-Direct substantially closes this gap through direct fine-tuning on full MITS data, and MILM-2S improves further, demonstrating that first learning from sampling patterns alone provides a better foundation.

Among the LLM-based irregular time series baselines, VITAL-LLM performs the weakest overall (avg.\ rank $13.1$), even trailing Qwen3-4B on several datasets. Both VITAL-LLM and ISTS-PLM require aligning numerical observations to the LLM's token embedding space via text prototype reprogramming in VITAL-LLM \cite{kwon2025mind}, and via learnable time, variable, and value embedders in ISTS-PLM \cite{zhang2025unleashing}. VITAL-LLM's reprogramming matrix spans the entire vocabulary, making it difficult to train. MILM sidesteps this alignment problem entirely by serializing MITS directly as XML-formatted text, preserving the LLM's pretrained token embeddings and the knowledge encoded within them, and outperforms both baselines across all four datasets.

Note that the benefit of the text channel varies noticeably between the two databases. MIMIC-IV contains rich free-text clinical notes authored by physicians, whereas eICU stores structured care plan entries generated through the Philips eCareManager telehealth system \cite{pollard2018eicu}, which are comparatively more templated and less semantically varied. This difference is reflected in a simple count: adding the text channel (``+ Note'') improves AU-ROC or AP over the numerical-only version in $13$ cases among unimodal baselines on MIMIC-IV, but in only $7$ on eICU.
Nevertheless, MILM-2S remains the top-ranked method on eICU as well, demonstrating that it can outperform baselines even when the text channel carries weaker predictive signal. Interestingly, the performance gap between MILM-2S and MILM-Direct is larger on eICU (e.g., $1.39$ AU-ROC on eICU-IHM) than on MIMIC-IV ($0.48$ on MIMIC-IV-IHM), suggesting that when the text channel carries weaker predictive signal, Stage 1's emphasis on sampling patterns may provide a more important initialization advantage.

\subsection{Value Redaction Evaluation} \label{subsec:value_redact}

\begin{figure}[t]
\centering
\begin{minipage}[c]{0.78\textwidth}
    \begin{subfigure}[b]{0.245\textwidth}
        \includegraphics[width=\textwidth]{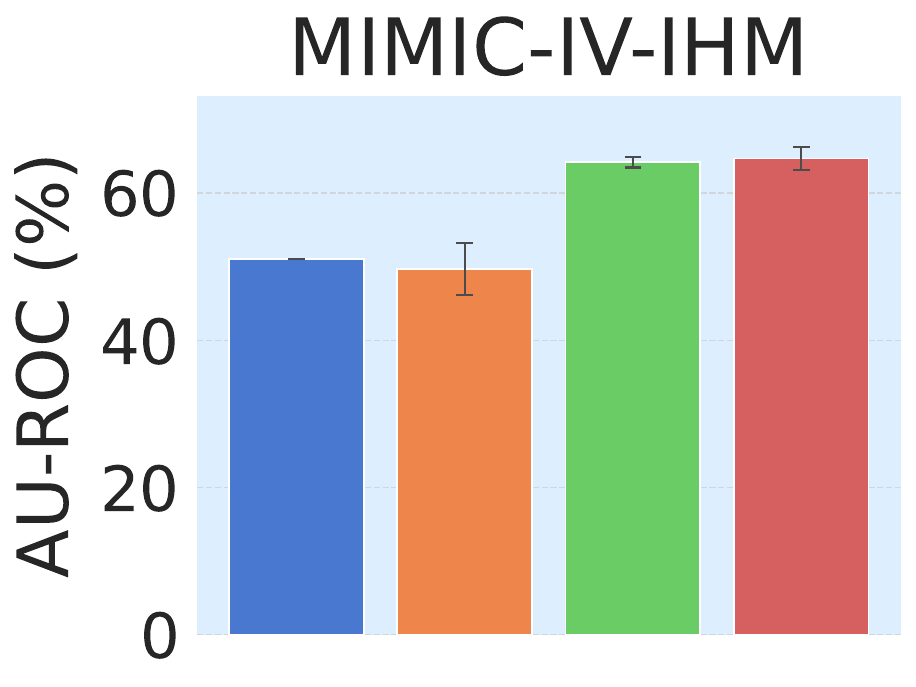}
    \end{subfigure}\hfill
    \begin{subfigure}[b]{0.24\textwidth}
        \includegraphics[width=\textwidth]{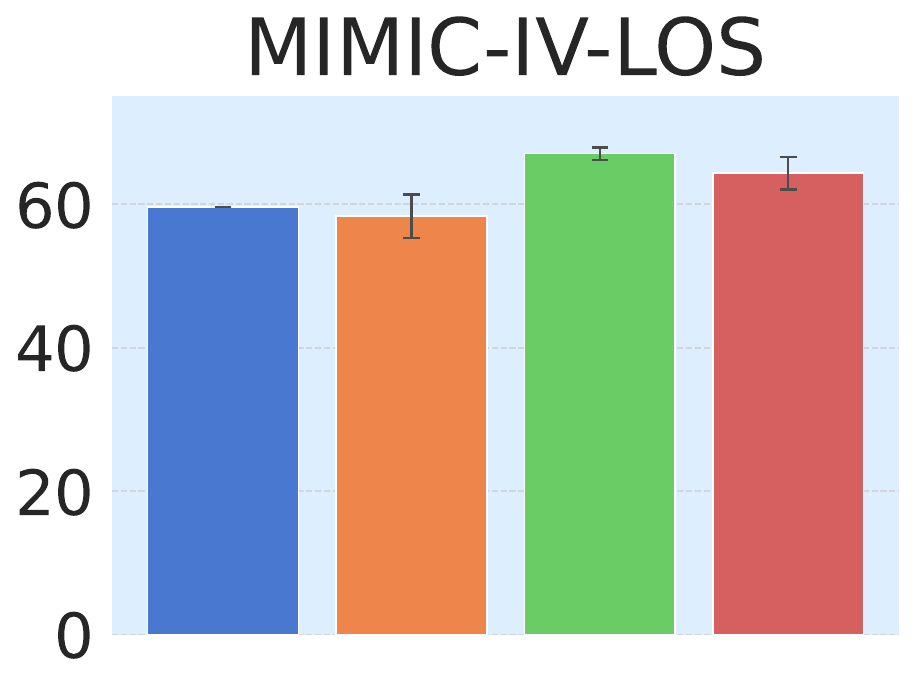}
    \end{subfigure}\hfill
    \begin{subfigure}[b]{0.24\textwidth}
        \includegraphics[width=\textwidth]{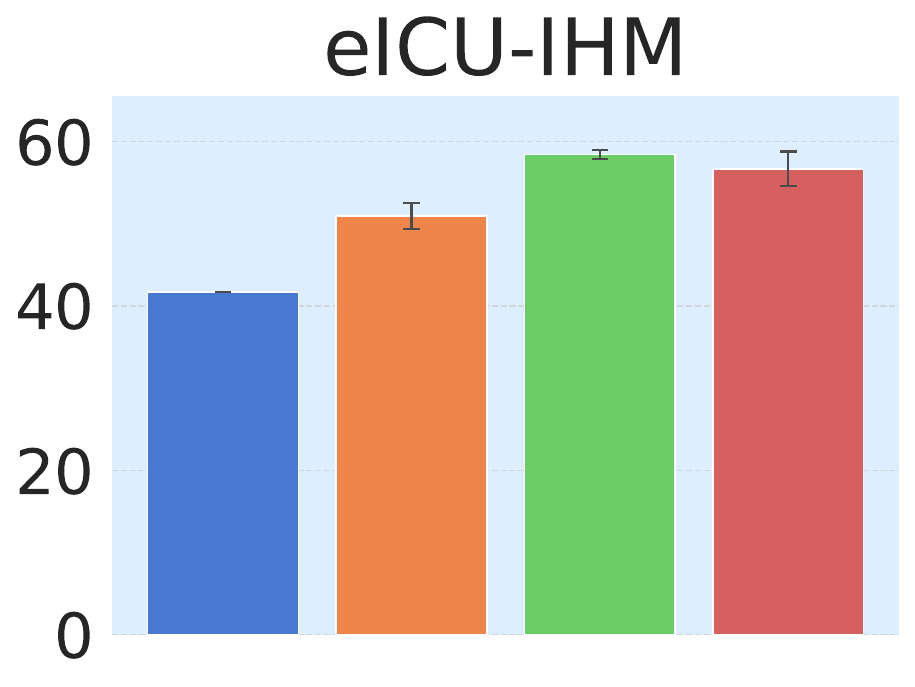}
    \end{subfigure}\hfill
    \begin{subfigure}[b]{0.24\textwidth}
        \includegraphics[width=\textwidth]{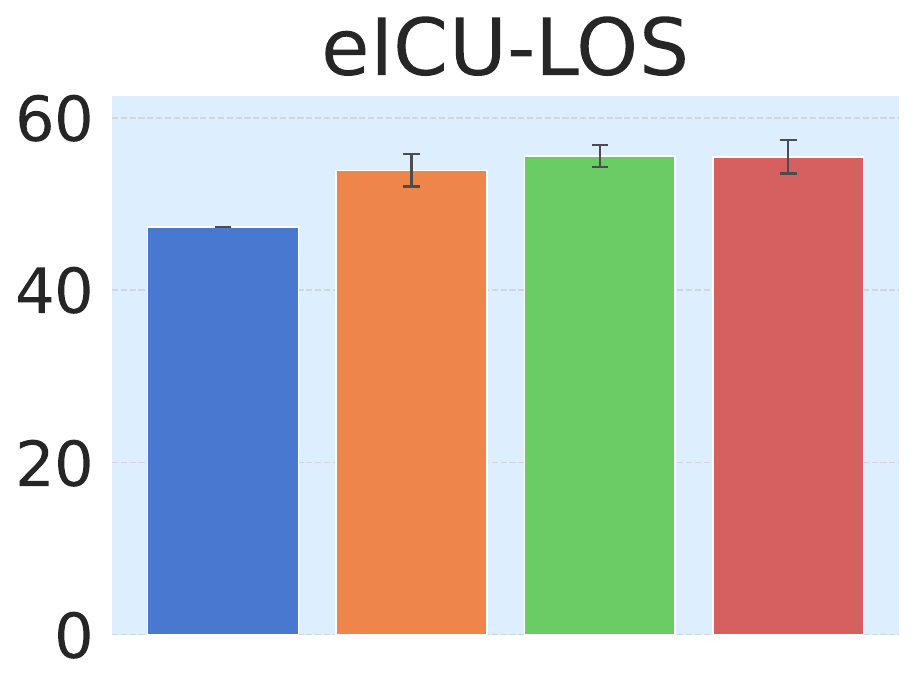}
    \end{subfigure}
    \\[0.5em]
    \begin{subfigure}[b]{0.245\textwidth}
        \includegraphics[width=\textwidth]{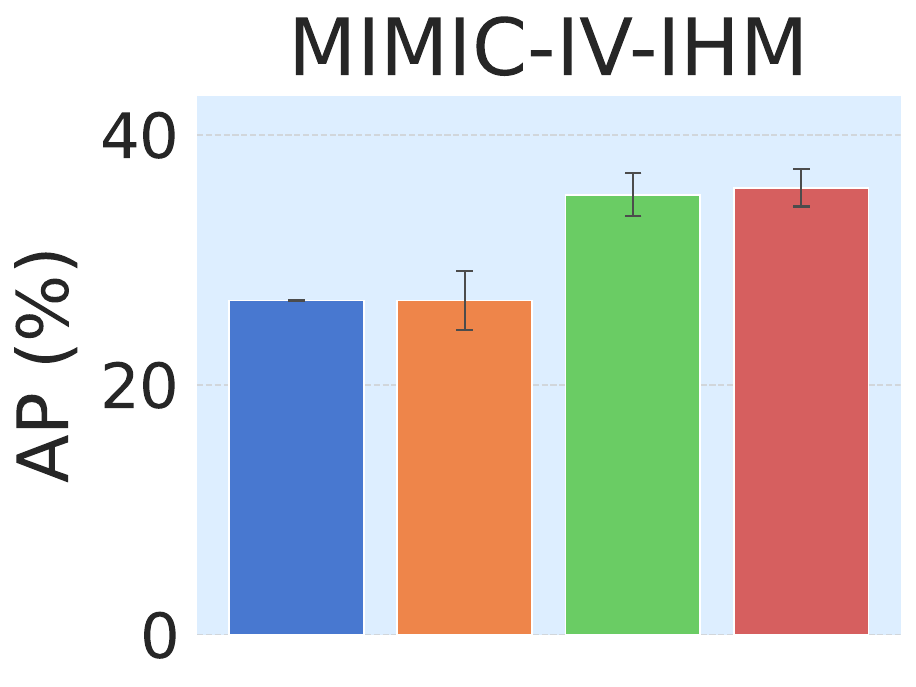}
    \end{subfigure}\hfill
    \begin{subfigure}[b]{0.24\textwidth}
        \includegraphics[width=\textwidth]{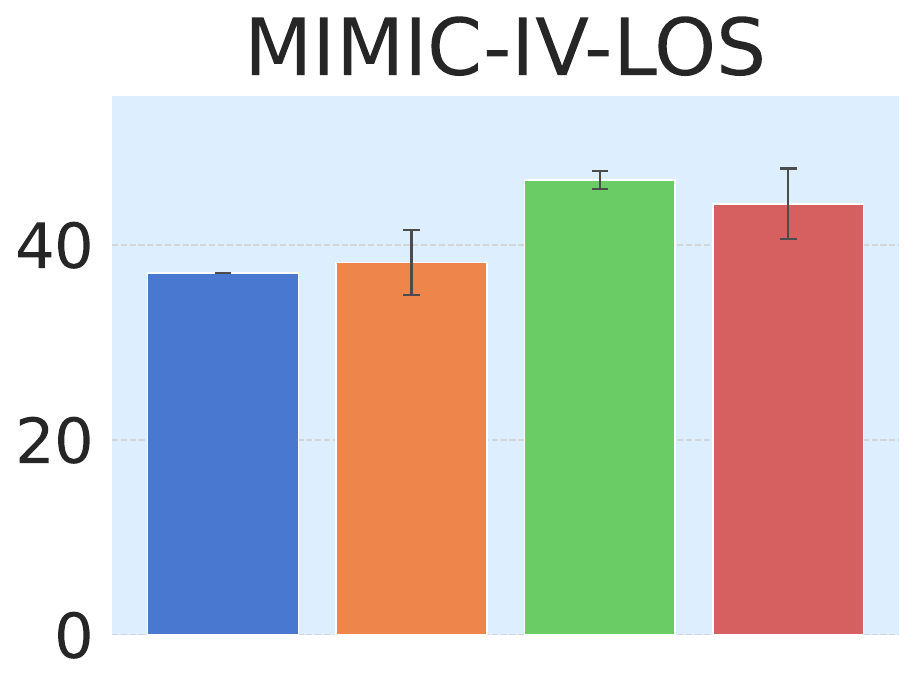}
    \end{subfigure}\hfill
    \begin{subfigure}[b]{0.24\textwidth}
        \includegraphics[width=\textwidth]{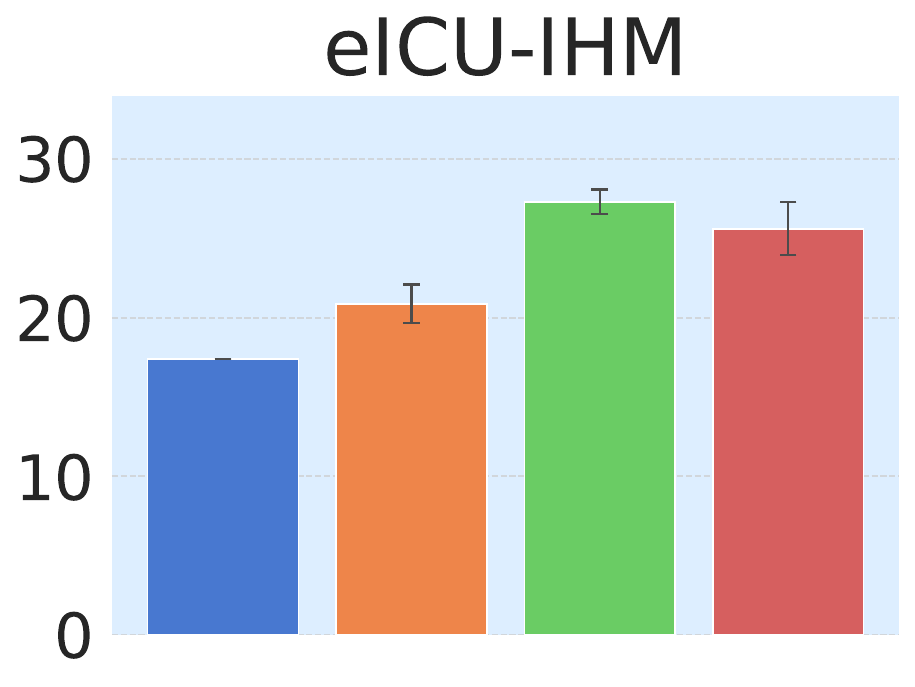}
    \end{subfigure}\hfill
    \begin{subfigure}[b]{0.24\textwidth}
        \includegraphics[width=\textwidth]{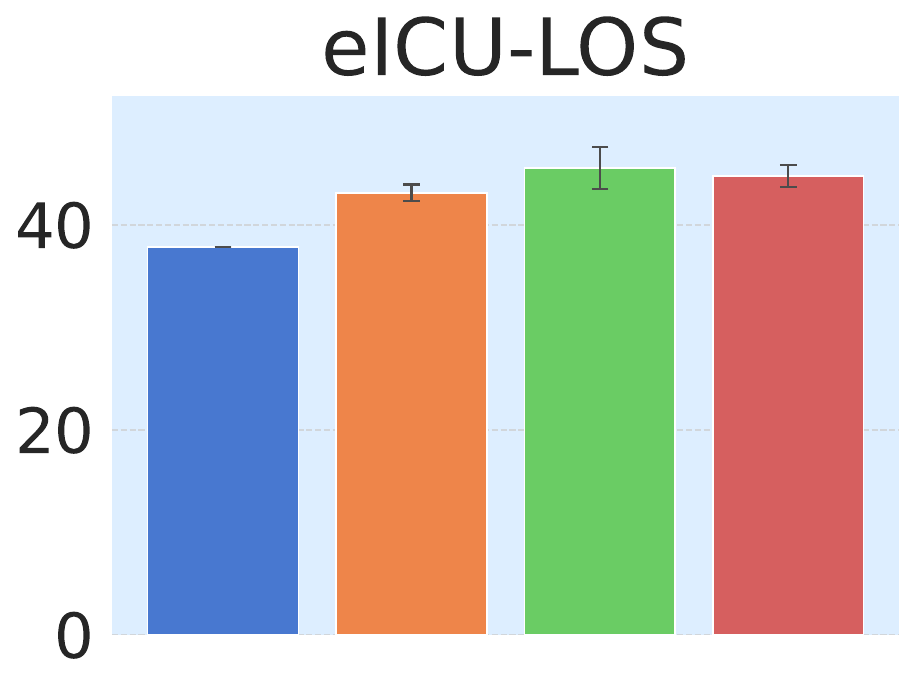}
    \end{subfigure}
\end{minipage}
\hfill
\begin{minipage}[c]{0.20\textwidth}
    \centering
    \includegraphics[width=\textwidth]{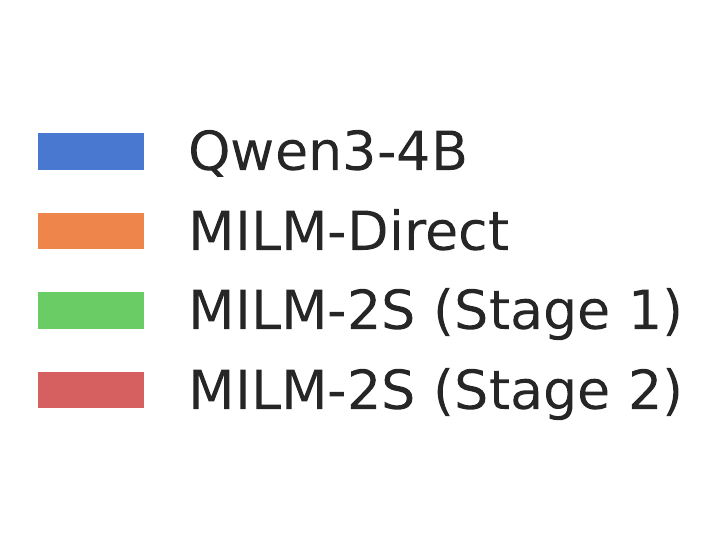}
\end{minipage}
\caption{Value redaction evaluation across datasets. All models are tested with observed values removed and only time and channel information retained. MILM-2S (Stage 1) substantially outperforms both Qwen3-4B and MILM-Direct, while MILM-2S (Stage 2) retains most of the performance.}
\label{fig:value_redaction}
\end{figure}

We conduct value redaction evaluation to understand how well each model exploits sampling patterns. We test off-the-shelf Qwen3-4B, MILM-Direct, and the Stage 1 and Stage 2 checkpoints of MILM-2S with all observed values removed, keeping only the time and channel information. The results are shown in Fig.~\ref{fig:value_redaction}. MILM-Direct does not improve over Qwen3-4B on MIMIC-IV, but does on eICU, where Qwen3-4B performs worse than random. MILM-2S (Stage 1) shows substantial improvement over both Qwen3-4B and MILM-Direct, demonstrating that sampling patterns are informative and can be learned through value redaction fine-tuning. MILM-2S (Stage 2) typically experiences a small degree of forgetting (e.g., a maximum decrease of $2.72$ AU-ROC on MIMIC-IV-LOS), with the exception of MIMIC-IV-IHM, where performance slightly increases. Nevertheless, MILM-2S (Stage 2) still predicts more accurately from sampling patterns alone than both Qwen3-4B and MILM-Direct, showing that it retains most sampling-pattern knowledge after full-data fine-tuning.

\subsection{Value Pending Evaluation}\label{subsec:value_pending}
\begin{table}[t]
\centering
\caption{Value pending evaluation results under two countermeasures: drop observation and show presence. Best results are \textbf{boldfaced} and second-best results are \underline{underlined}.}
\label{tab:value_pending}
\scalebox{0.9}{
\begin{tabular}{llcccc}
\toprule
& & \multicolumn{2}{c}{\textbf{MIMIC-IV-IHM}} & \multicolumn{2}{c}{\textbf{MIMIC-IV-LOS}} \\
\cmidrule(lr){3-4}
\cmidrule(lr){5-6}
\textbf{Countermeasure} & \textbf{Method} & AU-ROC & AP & AU-ROC & AP \\
\midrule
Drop Observation & MILM-Direct 
& \std{79.04}{0.91} & \std{59.23}{1.63} 
& \std{76.86}{0.52} & \std{56.44}{1.61} \\

& MILM-2S 
& \second{79.49}{0.76} & \second{60.02}{1.57} 
& \best{77.64}{0.33} & \best{57.17}{1.78} \\

\midrule
Show Presence & MILM-Direct 
& \std{79.21}{0.65} & \std{58.14}{1.37} 
& \std{76.30}{0.64} & \std{55.01}{2.07} \\

& MILM-2S 
& \best{80.34}{0.64} & \best{61.23}{1.62} 
& \second{77.08}{0.38} & \second{56.46}{1.50} \\
\bottomrule
\end{tabular}
}
\end{table}

We study a practical evaluation setup we call value pending evaluation. In practice, some lab measurements or notes are taken before the prediction time, but their results are not yet available in the database when the model is queried for a prediction \cite{otles2021mind}. For these observations, the time and channel information are known, but the values are not. We consider two countermeasures: dropping these observations entirely, or showing their presence (i.e., the fact that a channel was sampled at a particular time) without revealing the values. For example, in the following snippet:

\begin{lstlisting}[basicstyle=\scriptsize\ttfamily, columns=flexible, breaklines=false, frame=none,
backgroundcolor=\color{gray!10},
xleftmargin=12pt, xrightmargin=4pt,
aboveskip=3pt, belowskip=3pt]
<time> 19.10 hours </time> <channel> Anion Gap (mEq/L) </channel> <value> 14.00 </value>
<time> 19.10 hours </time> <channel> Bicarbonate (mEq/L) </channel>
<time> 19.10 hours </time> <channel> Red Blood Cells (K/uL) </channel> <value> 2.85 </value>
\end{lstlisting}

the Bicarbonate value measured at $19.10$ hours is not yet available, while those of Anion Gap and Red Blood Cells are. Showing presence conveys extra sampling information but risks exposing the model to a mixed format of pairs and triplets never seen before, challenging its generalization capability.

In MIMIC-IV, each lab event and clinical note has two associated timestamps: \inlinecode{charttime} and \inlinecode{storetime}. \inlinecode{charttime} is the clinically relevant event time used for MITS construction, e.g., time of specimen acquisition or note authoring, and \inlinecode{storetime} is when the result became available in the database. For the value pending evaluation, we withhold the values of observations whose \inlinecode{storetime} exceeds the prediction time of $24$ hours since ICU admission. eICU does not have an equivalent pair of time columns, so we focus this evaluation on MIMIC-IV. The results are shown in Table~\ref{tab:value_pending}. See Appendix~\ref{app:value_pending_eval} for dataset statistics and full results including baselines under drop observation.

Across both datasets and countermeasures, MILM-2S outperforms MILM-Direct by a larger margin compared to standard evaluation. With drop observation, MILM-2S better utilizes the remaining observations for prediction.
With show presence, MILM-2S handles the mixed pair-triplet format more effectively than MILM-Direct, as its two-stage training has exposed it to time-channel pairs in Stage 1 and full triplets in Stage 2, despite never training on the two formats mixed together.
On MIMIC-IV-LOS, show presence underperforms drop observation for MILM-2S, likely reflecting the slight forgetting of sampling-pattern knowledge during Stage 2 noted in Section~\ref{subsec:value_redact}. On MIMIC-IV-IHM, however, show presence achieves the best overall result, suggesting that MILM-2S can utilize the extra sampling information for better in-hospital mortality prediction.

Value-pending observations tend to fall at the end of the timeline and are often critical for predicting future outcomes \cite{choi2016retain}. Their unavailability can cause a nontrivial performance drop. We introduce value pending evaluation as a novel and practical evaluation setting, and encourage follow-up work in irregular time series modeling to consider this setting.



\section{Conclusion}

We introduce MILM, the first LLM-based framework for classifying multimodal irregular time series (MITS) with informative sampling. MILM represents observations as time-ordered triplets in XML format and applies a two-stage fine-tuning strategy. The first stage teaches the LLM to predict solely from sampling patterns, i.e., when and which channels are observed, while the second stage incorporates value information on top of the learned sampling patterns. The two-stage model achieves the strongest average performance across multiple EHR datasets. Value redaction evaluation further confirms that sampling patterns are predictive and that two-stage fine-tuning effectively learns and retains this predictive capability. We also introduce a practical value pending evaluation setup, where the values of some observations are unavailable at prediction time. In this setting, the two-stage model outperforms direct full data fine-tuning and leverages the time and channel information of value-pending observations to improve in-hospital mortality prediction. Future work includes extending MILM to additional modalities and studying robustness under sampling distribution shift, such as deployment across hospitals with different sampling policies.



\bibliographystyle{unsrtnat}   
\bibliography{refs}


\appendix

\section{Prompt Templates}\label{app:prompt}
We present the prompt templates for the in-hospital mortality (IHM) and length-of-stay (LOS) prediction tasks. The templates are organized into three groups based on the MITS representation used. The first group uses the full MITS representation, corresponding to Stage~2 of MILM and inference. The second group uses the value-redacted MITS representation, corresponding to Stage~1 of MILM. The third group corresponds to value pending evaluation with show presence, where some observations have been measured, but their values are unavailable at prediction time and are shown as time-channel pairs. 

\subsection{Prompt Templates for Full MITS}

\begin{tcolorbox}[
    enhanced,
    breakable,
    colback=teal!6,                          
    colframe=teal!40,                        
    arc=3pt,
    boxrule=0.7pt,
    title={\small\bfseries Full MITS Prompt Template: In-Hospital Mortality Prediction},
    coltitle=white,
    colbacktitle=teal!50!black,              
    toptitle=2pt, bottomtitle=2pt,
    top=5pt, bottom=5pt, left=6pt, right=6pt,
    title after break={\small\bfseries Prompt Template (continued)},
    extras broken={colbacktitle=teal!50!black},  
]
\setlength{\parindent}{0pt}%
\setlength{\parskip}{0pt}%
%
\colorbox{teal!12}{\footnotesize\bfseries\sffamily System}\par   
\vspace{3pt}
{\scriptsize\ttfamily                             
You are a critical care physician predicting outcomes from ICU monitoring data including laboratory measurements and radiology notes.
}\\
\vspace{3pt}\noindent\rule{\linewidth}{0.3pt}\vspace{4pt}
%
\colorbox{teal!12}{\footnotesize\bfseries\sffamily User}\par     
\vspace{3pt}
{\scriptsize\ttfamily                             
TASK: Analyze timestamped observations from an ICU stay and predict whether the
patient will survive their hospital admission.\\
\\
DATA FORMAT: Each observation is a triplet:\\
\hspace*{1.5em}<time> hours\_since\_start </time> <channel> channel\_name
</channel> <value> value </value>\\
\\
OUTPUT FORMAT (FOLLOW EXACTLY):\\
Provide your answer in <answer> </answer> tags.\\
Write ONLY a single letter: A or B\\
\\
(A) SURVIVAL\\
(B) MORTALITY\\
\\
Do NOT include any explanatory text outside the tags.
Start directly with <answer>.\\
\\
Example 1 (SURVIVAL): <answer> A </answer>\\
Example 2 (MORTALITY): <answer> B </answer>\\
\\
{[OBSERVATIONS]}\\
\ldots \\
{[/OBSERVATIONS]}\\
\\
Based on the observations above, predict whether this patient will experience
in-hospital mortality.\\
Follow the output format exactly. Start directly with <answer>.
}
\end{tcolorbox}

\begin{tcolorbox}[
    enhanced,
    breakable,
    colback=teal!6,
    colframe=teal!40,
    arc=3pt,
    boxrule=0.7pt,
    title={\small\bfseries Full MITS Prompt Template: Length-of-Stay Prediction},
    coltitle=white,
    colbacktitle=teal!50!black,
    toptitle=2pt, bottomtitle=2pt,
    top=5pt, bottom=5pt, left=6pt, right=6pt,
    title after break={\small\bfseries Prompt Template (continued)},
    extras broken={colbacktitle=teal!50!black},
]
\setlength{\parindent}{0pt}%
\setlength{\parskip}{0pt}%
%
\colorbox{teal!12}{\footnotesize\bfseries\sffamily System}\par
\vspace{3pt}
{\scriptsize\ttfamily
You are a critical care physician predicting outcomes from ICU monitoring data including laboratory measurements and radiology notes.
}\\
\vspace{3pt}\noindent\rule{\linewidth}{0.3pt}\vspace{4pt}
%
\colorbox{teal!12}{\footnotesize\bfseries\sffamily User}\par
\vspace{3pt}
{\scriptsize\ttfamily
TASK: Given timestamped observations from the first 24-hour window of an ICU
stay, predict whether the patient will ultimately have a short ICU stay
(<96 hours) and survive.\\        
\\
DATA FORMAT: Each observation is a triplet:\\
\hspace*{1.5em}<time> hours\_since\_start </time> <channel> channel\_name
</channel> <value> value </value>\\
\\
CLASS DEFINITION:\\               
- SHORT\_STAY: ICU stay < 96 hours and survived.\\
- LONG\_STAY: ICU stay >= 96 hours or death.\\
\\
OUTPUT FORMAT (FOLLOW EXACTLY):\\
Provide your answer in <answer> </answer> tags.\\
Write ONLY a single letter: A or B\\
\\
(A) LONG\_STAY\\                  
(B) SHORT\_STAY\\                 
\\
Do NOT include any explanatory text outside the tags.
Start directly with <answer>.\\
\\
Example 1 (LONG\_STAY): <answer> A </answer>\\    
Example 2 (SHORT\_STAY): <answer> B </answer>\\   
\\
{[OBSERVATIONS]}\\
\ldots \\
{[/OBSERVATIONS]}\\
\\
Based on the observations above, predict whether this patient will have a
short ICU stay and survive.\\     
Follow the output format exactly. Start directly with <answer>.
}
\end{tcolorbox}

\subsection{Prompt Templates for Value-Redacted MITS}

\begin{tcolorbox}[
    enhanced,
    breakable,
    colback=teal!6,                          
    colframe=teal!40,                        
    arc=3pt,
    boxrule=0.7pt,
    title={\small\bfseries Value-Redacted MITS Prompt Template: In-Hospital Mortality Prediction},
    coltitle=white,
    colbacktitle=teal!50!black,              
    toptitle=2pt, bottomtitle=2pt,
    top=5pt, bottom=5pt, left=6pt, right=6pt,
    title after break={\small\bfseries Prompt Template (continued)},
    extras broken={colbacktitle=teal!50!black},  
]
\setlength{\parindent}{0pt}%
\setlength{\parskip}{0pt}%
%
\colorbox{teal!12}{\footnotesize\bfseries\sffamily System}\par   
\vspace{3pt}
{\scriptsize\ttfamily                             
You are a critical care physician predicting outcomes from ICU monitoring data including laboratory measurements and radiology notes.
}\\
\vspace{3pt}\noindent\rule{\linewidth}{0.3pt}\vspace{4pt}
%
\colorbox{teal!12}{\footnotesize\bfseries\sffamily User}\par     
\vspace{3pt}
{\scriptsize\ttfamily                             
TASK: Analyze timestamped observations from an ICU stay and predict whether the patient will survive their hospital admission.\\
\\
DATA FORMAT: Each observation is a pair:\\
<time> hours\_since\_start </time> <channel> channel\_name </channel>\\
\\
IMPORTANT:\\
- For each observation, the value/content was measured and recorded, but is not provided here.\\
- When making the prediction, use the information conveyed by the timings and channel types themselves — what types of laboratory measurements were ordered and what notes were recorded, when, in what sequence, and how frequently — and what these patterns might signal about the patient's clinical trajectory and mortality risk.\\
\\
OUTPUT FORMAT (FOLLOW EXACTLY):\\
Provide your answer in <answer> </answer> tags.\\
Write ONLY a single letter: A or B\\
\\
(A) SURVIVAL\\
(B) MORTALITY\\
\\
Do NOT include any explanatory text outside the tags. Start directly with <answer>.\\
\\
Example 1 (SURVIVAL): <answer> A </answer>\\
Example 2 (MORTALITY): <answer> B </answer>\\
\\
{[OBSERVATIONS]}\\
\ldots\\
{[/OBSERVATIONS]}\\
\\
Based on the observations above, predict whether this patient will experience in-hospital mortality.\\
Follow the output format exactly. Start directly with <answer>.
}
\end{tcolorbox}

\begin{tcolorbox}[
    enhanced,
    breakable,
    colback=teal!6,
    colframe=teal!40,
    arc=3pt,
    boxrule=0.7pt,
    title={\small\bfseries Value-Redacted MITS Prompt Template: Length-of-Stay Prediction},
    coltitle=white,
    colbacktitle=teal!50!black,
    toptitle=2pt, bottomtitle=2pt,
    top=5pt, bottom=5pt, left=6pt, right=6pt,
    title after break={\small\bfseries Prompt Template (continued)},
    extras broken={colbacktitle=teal!50!black},
]
\setlength{\parindent}{0pt}%
\setlength{\parskip}{0pt}%
%
\colorbox{teal!12}{\footnotesize\bfseries\sffamily System}\par
\vspace{3pt}
{\scriptsize\ttfamily
You are a critical care physician predicting outcomes from ICU monitoring data including laboratory measurements and radiology notes.
}\\
\vspace{3pt}\noindent\rule{\linewidth}{0.3pt}\vspace{4pt}
%
\colorbox{teal!12}{\footnotesize\bfseries\sffamily User}\par
\vspace{3pt}
{\scriptsize\ttfamily
TASK: Given timestamped observations from the first 24-hour window of an ICU stay, predict whether the patient will ultimately have a short ICU stay (<96 hours) and survive.\\
\\
DATA FORMAT: Each observation is a pair:\\
<time> hours\_since\_start </time> <channel> channel\_name </channel>\\
\\
CLASS DEFINITION:\\
- SHORT\_STAY: ICU stay < 96 hours and survived.\\
- LONG\_STAY: ICU stay >= 96 hours or death.\\
\\
IMPORTANT:\\
- For each observation, the value/content was measured and recorded, but is not provided here.\\
- When making the prediction, use the information conveyed by the timings and channel types themselves - what types of laboratory measurements were ordered and what notes were recorded, when, in what sequence, and how frequently - and what these patterns might signal about whether the patient's ICU course is likely to stabilize quickly (short stay and survive) versus become prolonged or deteriorate (ICU stay >= 96 hours or death).\\
\\
OUTPUT FORMAT (FOLLOW EXACTLY):\\
Provide your answer in <answer> </answer> tags.\\
Write ONLY a single letter: A or B\\
\\
(A) LONG\_STAY\\
(B) SHORT\_STAY\\
\\
Do NOT include any explanatory text outside the tags. Start directly with <answer>.\\
\\
Example 1 (LONG\_STAY): <answer> A </answer>\\
Example 2 (SHORT\_STAY): <answer> B </answer>\\
\\
{[OBSERVATIONS]}\\
\ldots\\
{[/OBSERVATIONS]}\\
\\
Based on the observations above, predict whether this patient will have a short ICU stay and survive.\\
Follow the output format exactly. Start directly with <answer>.
}
\end{tcolorbox}

\subsection{Prompt Templates for Value Pending Evaluation with Show Presence}

\begin{tcolorbox}[
    enhanced,
    breakable,
    colback=teal!6,                          
    colframe=teal!40,                        
    arc=3pt,
    boxrule=0.7pt,
    title={\small\bfseries Value Pending Evaluation with Show Presence Prompt Template: In-Hospital Mortality Prediction},
    coltitle=white,
    colbacktitle=teal!50!black,              
    toptitle=2pt, bottomtitle=2pt,
    top=5pt, bottom=5pt, left=6pt, right=6pt,
    title after break={\small\bfseries Prompt Template (continued)},
    extras broken={colbacktitle=teal!50!black},  
]
\setlength{\parindent}{0pt}%
\setlength{\parskip}{0pt}%
%
\colorbox{teal!12}{\footnotesize\bfseries\sffamily System}\par   
\vspace{3pt}
{\scriptsize\ttfamily                             
You are a critical care physician predicting outcomes from ICU monitoring data including laboratory measurements and radiology notes.
}\\
\vspace{3pt}\noindent\rule{\linewidth}{0.3pt}\vspace{4pt}
%
\colorbox{teal!12}{\footnotesize\bfseries\sffamily User}\par     
\vspace{3pt}
{\scriptsize\ttfamily                             
TASK: Analyze timestamped observations from an ICU stay and predict whether the patient will survive their hospital admission.\\
\\
DATA FORMAT:\\
Each observation is one of the following:\\
1. A triplet:\\
<time> hours\_since\_start </time> <channel> channel\_name </channel> <value> value </value>\\

2. A pair:\\
<time> hours\_since\_start </time> <channel> channel\_name </channel>\\

IMPORTANT:\\
- For some observations, the measurement was made but the value/content is not available at the time of prediction. These observations are shown as time-channel pairs.\\
- When making the prediction, use both the available values/content and the information conveyed by the timings and channel types themselves — what types of laboratory measurements were ordered and what notes were recorded, when, in what sequence, and how frequently — and what these patterns might signal about the patient's clinical trajectory and mortality risk.\\
\\
OUTPUT FORMAT (FOLLOW EXACTLY):\\
Provide your answer in <answer> </answer> tags.\\
Write ONLY a single letter: A or B\\
\\
(A) SURVIVAL\\
(B) MORTALITY\\
\\
Do NOT include any explanatory text outside the tags. Start directly with <answer>.\\
\\
Example 1 (SURVIVAL): <answer> A </answer>\\
Example 2 (MORTALITY): <answer> B </answer>\\
\\
{[OBSERVATIONS]}\\
\ldots\\
{[/OBSERVATIONS]}\\
\\
Based on the observations above, predict whether this patient will experience in-hospital mortality.\\
Follow the output format exactly. Start directly with <answer>.
}
\end{tcolorbox}

\begin{tcolorbox}[
    enhanced,
    breakable,
    colback=teal!6,
    colframe=teal!40,
    arc=3pt,
    boxrule=0.7pt,
    title={\small\bfseries Value Pending with Show Presence Evaluation Prompt Template: Length-of-Stay Prediction},
    coltitle=white,
    colbacktitle=teal!50!black,
    toptitle=2pt, bottomtitle=2pt,
    top=5pt, bottom=5pt, left=6pt, right=6pt,
    title after break={\small\bfseries Prompt Template (continued)},
    extras broken={colbacktitle=teal!50!black},
]
\setlength{\parindent}{0pt}%
\setlength{\parskip}{0pt}%
%
\colorbox{teal!12}{\footnotesize\bfseries\sffamily System}\par
\vspace{3pt}
{\scriptsize\ttfamily
You are a critical care physician predicting outcomes from ICU monitoring data including laboratory measurements and radiology notes.
}\\
\vspace{3pt}\noindent\rule{\linewidth}{0.3pt}\vspace{4pt}
%
\colorbox{teal!12}{\footnotesize\bfseries\sffamily User}\par
\vspace{3pt}
{\scriptsize\ttfamily
TASK: Given timestamped observations from the first 24-hour window of an ICU stay, predict whether the patient will ultimately have a short ICU stay (<96 hours) and survive.\\
\\
DATA FORMAT:\\
Each observation is one of the following:\\
1. A triplet:\\
<time> hours\_since\_start </time> <channel> channel\_name </channel> <value> value </value>\\
\\
2. A pair:\\
<time> hours\_since\_start </time> <channel> channel\_name </channel>\\
\\
CLASS DEFINITION:\\
- SHORT\_STAY: ICU stay < 96 hours and survived.\\
- LONG\_STAY: ICU stay >= 96 hours or death.\\
\\
IMPORTANT:\\
- For some observations, the measurement was made but the value/content is not available at the time of prediction. These observations are shown as time-channel pairs.\\
- When making the prediction, use both the available values/content and the information conveyed by the timings and channel types themselves - what types of laboratory measurements were ordered and what notes were recorded, when, in what sequence, and how frequently - and what these patterns might signal about whether the patient's ICU course is likely to stabilize quickly (short stay and survive) versus become prolonged or deteriorate (ICU stay >= 96 hours or death).\\
\\
OUTPUT FORMAT (FOLLOW EXACTLY):\\
Provide your answer in <answer> </answer> tags.\\
Write ONLY a single letter: A or B\\
\\
(A) LONG\_STAY\\
(B) SHORT\_STAY\\
\\
Do NOT include any explanatory text outside the tags. Start directly with <answer>.\\
\\
Example 1 (LONG\_STAY): <answer> A </answer>\\
Example 2 (SHORT\_STAY): <answer> B </answer>\\
\\
{[OBSERVATIONS]}\\
\ldots\\
{[/OBSERVATIONS]}\\
\\
Based on the observations above, predict whether this patient will have a short ICU stay and survive.\\
Follow the output format exactly. Start directly with <answer>.
}
\end{tcolorbox}

\section{Dataset Details}\label{app:dataset}
\subsection{EHR Database Overview}
\paragraph{MIMIC-IV.} The Medical Information Mart for Intensive Care IV (MIMIC-IV)~\cite{johnson2023mimic,PhysioNet-mimiciv-3.1,PhysioNet-mimic-iv-note-2.2} is a publicly available EHR dataset sourced from the Beth Israel Deaconess Medical Center (BIDMC) in Boston, MA, covering hospital and ICU admissions from 2008 to 2022. MIMIC-IV is organized into three modules. The \inlinecode{hosp} module contains hospital-wide data, including laboratory measurements, microbiology cultures, medication orders, billing, and patient tracking. The \inlinecode{icu} module contains ICU bedside data, including charted observations, infusions, and patient outputs, recorded with both a clinically relevant timestamp (\inlinecode{charttime}) and a database availability timestamp (\inlinecode{storetime}). The \inlinecode{note} module contains deidentified free-text clinical notes, including discharge summaries and radiology reports authored by physicians. Data were deidentified in accordance with the HIPAA Safe Harbor provision and are available via PhysioNet.
\paragraph{eICU.} The eICU Collaborative Research Database (eICU-CRD) v2.0 ~\cite{pollard2018eicu,PhysioNet-eicu-crd-2.0} is a publicly available multi-center critical care database built from data collected through the Philips eICU Program, a telehealth ICU monitoring system in which remote care teams monitor patients at participating hospitals via the eCareManager platform. The database spans admissions from 2014 to 2015 across hundreds of ICU units at hospitals throughout the United States. Available data include vital signs, laboratory measurements, medications, APACHE severity scores, admission diagnoses, and care plan documentation. The care plan is documented using structured multiple-choice lists covering care provider types, prognosis, treatment status, goals of care, and end-of-life planning. Unlike the rich physician-authored free-text notes in MIMIC-IV, eICU care plan entries are comparatively templated and less linguistically varied. The database is deidentified per HIPAA Safe Harbor and available via PhysioNet.

\subsection{Dataset Extraction}
We describe the extraction pipeline for each database separately, covering cohort definition, channel selection, filtering criteria, and label construction. Both databases use a 24-hour observation window from ICU admission and a 70\%/15\%/15\% random train/validation/test split at the ICU stay level.

\subsubsection{MIMIC-IV}

\paragraph{Numerical Channels.} We extract $21$ laboratory tests from the \inlinecode{labevents} table in the MIMIC-IV \inlinecode{hosp} module: Glucose, Sodium, Chloride, Creatinine, Urea Nitrogen, Bicarbonate, Anion Gap, Hemoglobin, Hematocrit, Magnesium, Platelet Count, Phosphate, White Blood Cells, Calcium (Total), MCH, Red Blood Cells, MCHC, MCV, RDW, Neutrophils, and Vancomycin. Each lab measurement is timestamped by its \inlinecode{charttime}, which records the clinically relevant time of specimen collection. The \inlinecode{storetime}, which records when the result became available in the database, is also retained for the value pending evaluation in Section~\ref{subsec:value_pending}.

\paragraph{Text Channel.} We extract radiology notes from MIMIC-IV-Note ~\cite{PhysioNet-mimic-iv-note-2.2}, which contains deidentified free-text radiology reports authored by physicians. Each note carries a \inlinecode{charttime} and a \inlinecode{storetime}. Discharge summaries are excluded as they may contain explicit outcome information. Radiology reports do not have this concern.

\paragraph{Cohort and Filtering.} For both tasks, we restrict observations to the first 24 hours of the ICU stay. We require each stay to have at least $64$ lab measurements and between 2 and 6 notes within this window, and retain only stays satisfying both modality requirements. The lower bound on note count ensures that notes form a temporal series rather than an isolated observation, while the upper bound removes ICU stays with an anomalously high number of notes to prevent context length explosion.

\paragraph{IHM Task.} The cohort consists of ICU stays with a length of stay (LOS) of at least 24 hours. The label is in-hospital mortality, derived from the \inlinecode{hospital\_expire\_flag} field in the admissions table.

\paragraph{LOS Task.} The cohort consists of ICU stays with an LOS of at least $48$ hours. Following~\citet{han2024fusemoe}, the binary label is short stay, defined as an ICU stay shorter than $96$ hours with the patient surviving, and $0$ otherwise.

\subsubsection{eICU}

\paragraph{Numerical Channels.} We extract $24$ laboratory tests from the lab table: glucose, potassium, sodium, chloride, creatinine, BUN, bicarbonate, anion gap, Hgb, Hct, magnesium, platelets, phosphate, WBC, calcium, MCH, RBC, MCHC, MCV, RDW, neutrophils, and Vancomycin (random, trough, and peak). These correspond to the same clinical measurements selected for MIMIC-IV. Timestamps are derived from \inlinecode{labresultoffset}, which records minutes elapsed since admission and is converted to hours.

\paragraph{Text Channel.} We extract care plan notes from the \inlinecode{note} table, which stores clinical documentation entered through the eCareManager telehealth platform. Each row in the table represents a single form field selection, with \inlinecode{notepath} encoding a full hierarchy of section, field label, and field value (e.g., \inlinecode{notes/Progress Notes/Interventions/Major/Sepsis -\ evaluation and management}), and \inlinecode{notevalue} storing the authoritative leaf value. We define a note event as all rows sharing the same patient stay, timestamp (\inlinecode{noteoffset}), and note type (e.g., \inlinecode{Brief Progress}, \inlinecode{Admission}, \inlinecode{Intubation}), and exclude CPR-type notes, as their presence may constitute label leakage for the mortality prediction task.

Because the eCareManager system interleaves clinical documentation with UI control fields (e.g., print settings, view mode toggles, sign-off confirmations), many rows in the table are structural artifacts rather than clinical observations. We filter out these noise rows, specifically those whose section corresponds to system UI paths (e.g., \inlinecode{View and Save}), whose field label corresponds to scaffolding controls (e.g., \inlinecode{View Options}, \inlinecode{Save Options}, \inlinecode{Sign As}), or whose field value is a generic modal selection (e.g., \inlinecode{System View}, \inlinecode{Performed}, \inlinecode{Not Significant}).

For the remaining informative rows, we format each field differently depending on its section. In the \inlinecode{Interventions} and \inlinecode{Assessment and Plan} sections, the field value itself is the full clinical problem or intervention name (e.g., \inlinecode{Respiratory failure - evaluation and management}), so we emit only the value. In all other sections, the field label names a clinical attribute and the value records the finding, so we emit a \inlinecode{field\_label: field\_value} token (e.g., \inlinecode{Smoking Status: denies smoking}). All informative tokens from a note event are then deduplicated and concatenated into a single note string prefixed with its note type. For example, a \inlinecode{Brief Progress} note event might produce:

\begin{lstlisting}[basicstyle=\scriptsize\ttfamily, columns=flexible, breaklines=false, frame=none,
backgroundcolor=\color{gray!10},
xleftmargin=12pt, xrightmargin=4pt,
aboveskip=3pt, belowskip=3pt]
[Brief Progress]
Communication with other healthcare providers and/or family
Respiratory failure - evaluation and management
\end{lstlisting}

and an \inlinecode{Admission} note event might produce:

\begin{lstlisting}[basicstyle=\scriptsize\ttfamily, columns=flexible, breaklines=false, frame=none,
backgroundcolor=\color{gray!10},
xleftmargin=12pt, xrightmargin=4pt,
aboveskip=3pt, belowskip=3pt]
[Admission]
Smoking Status: denies smoking
Ethanol Use: rare
\end{lstlisting}

Note events whose reconstructed text is empty after noise removal are discarded.

\paragraph{Cohort and Filtering.} We restrict observations to the first 24 hours of the ICU stay. We require each stay to have at least $16$ lab measurements and between 3 and 10 care plan note events within this window, and retain only stays satisfying both modality requirements.

\paragraph{IHM Task.} The cohort consists of ICU stays with a LOS of at least 24 hours. The binary label is in-hospital mortality, defined as \inlinecode{hospitaldischargestatus == "Expired"}.

\paragraph{LOS Task.} The cohort consists of ICU stays with a LOS of at least 48 hours. The binary label is short stay, defined as an ICU stay shorter than 96 hours with the patient surviving at unit discharge (\inlinecode{unitdischargestatus == "Alive"}).

\subsection{Dataset Statistics and Analyses}

\begin{table}[h]
\centering
\caption{%
  Dataset statistics.
}
\label{tab:dataset_stats}
\scalebox{0.9}{
\begin{tabular}{lcccc}
\toprule
& \textbf{MIMIC-IV-IHM} & \textbf{MIMIC-IV-LOS} & \textbf{eICU-IHM} & \textbf{eICU-LOS} \\
\midrule
\# MITS Samples      & 3,404              & 2,896              & 3,667              & 2,712              \\
Positive Rate (\%)   & 25.79              & 27.83              & 22.42              & 37.13               \\
\# Labs Per MITS     & \std{81.9}{17.7}   & \std{82.2}{17.8}   & \std{36.4}{18.8}   & \std{36.8}{19.0}    \\
\# Notes Per MITS    & \std{3.12}{1.17}   & \std{3.15}{1.17}   & \std{3.86}{1.23}   & \std{3.87}{1.24}    \\
\bottomrule
\end{tabular}
}
\end{table}

\begin{table}[t]
\centering
\caption{%
  Lab measurement counts per sample per channel. For eICU, Vancomycin ($^{*}$) aggregates the random, trough, and peak sub-types.
}
\label{tab:lab_channel_stats}
\scalebox{0.9}{
\begin{tabular}{lcccc}
\toprule
 & \multicolumn{2}{c}{\textbf{MIMIC-IV}} & \multicolumn{2}{c}{\textbf{eICU}} \\
\cmidrule(lr){2-3}\cmidrule(lr){4-5}
\textbf{Channel} & \textbf{IHM} & \textbf{LOS} & \textbf{IHM} & \textbf{LOS} \\
\midrule
Anion Gap                                     & \std{4.18}{1.39} & \std{4.17}{1.39} & \std{1.73}{1.37} & \std{1.75}{1.39} \\
Bicarbonate                               & \std{4.21}{1.38} & \std{4.20}{1.37} & \std{1.99}{1.28} & \std{2.02}{1.30} \\
Blood Urea Nitrogen (BUN)                         & \std{4.22}{1.35} & \std{4.21}{1.35} & \std{1.98}{1.27} & \std{2.00}{1.27} \\
Calcium, Total                                & \std{3.78}{1.74} & \std{3.76}{1.73} & \std{1.93}{1.27} & \std{1.95}{1.28} \\
Chloride                                      & \std{4.33}{1.36} & \std{4.32}{1.36} & \std{2.01}{1.29} & \std{2.04}{1.30} \\
Creatinine                                    & \std{4.22}{1.36} & \std{4.21}{1.35} & \std{1.98}{1.26} & \std{2.00}{1.26} \\
Glucose                                      & \std{3.43}{4.35} & \std{3.57}{4.46} & \std{2.03}{1.36} & \std{2.06}{1.39} \\
Hematocrit (Hct)                                  & \std{5.28}{1.70} & \std{5.31}{1.74} & \std{1.90}{1.30} & \std{1.93}{1.30} \\
Hemoglobin (Hgb)                                  & \std{1.91}{2.59} & \std{2.00}{2.68} & \std{1.94}{1.35} & \std{1.95}{1.32} \\
Magnesium                                     & \std{3.98}{1.55} & \std{3.97}{1.54} & \std{1.29}{1.17} & \std{1.33}{1.19} \\
MCH                 & \std{4.73}{1.49} & \std{4.75}{1.51} & \std{1.64}{1.01} & \std{1.66}{1.00} \\
MCHC  & \std{4.74}{1.49} & \std{4.75}{1.51} & \std{1.65}{1.01} & \std{1.66}{1.00} \\
MCV                     & \std{4.74}{1.49} & \std{4.75}{1.50} & \std{1.65}{1.01} & \std{1.66}{1.01} \\
Neutrophils (-polys)                              & \std{0.77}{0.89} & \std{0.79}{0.90} & \std{0.69}{0.83} & \std{0.70}{0.85} \\
Phosphate                                   & \std{3.80}{1.72} & \std{3.78}{1.72} & \std{0.97}{1.13} & \std{1.00}{1.14} \\
Platelet Count                               & \std{4.96}{1.70} & \std{4.98}{1.73} & \std{1.67}{1.03} & \std{1.68}{1.02} \\
Potassium                          & ---              & ---              & \std{2.26}{1.49} & \std{2.28}{1.49} \\
Red Blood Cells (RBC)                             & \std{4.74}{1.49} & \std{4.75}{1.50} & \std{1.65}{1.00} & \std{1.67}{1.00} \\
RDW                 & \std{4.73}{1.49} & \std{4.75}{1.51} & \std{1.50}{1.04} & \std{1.53}{1.05} \\
Sodium                                       & \std{4.27}{1.47} & \std{4.27}{1.47} & \std{2.17}{1.52} & \std{2.19}{1.51} \\
Vancomycin$^{*}$                           & \std{0.13}{0.38} & \std{0.13}{0.38} & \std{0.07}{0.28} & \std{0.07}{0.28} \\
White Blood Cells (WBC)                           & \std{4.74}{1.49} & \std{4.76}{1.51} & \std{1.65}{1.00} & \std{1.67}{0.99} \\
\bottomrule
\end{tabular}
}
\end{table}

\begin{figure}[t]
\centering
\includegraphics[width=0.26\textwidth]{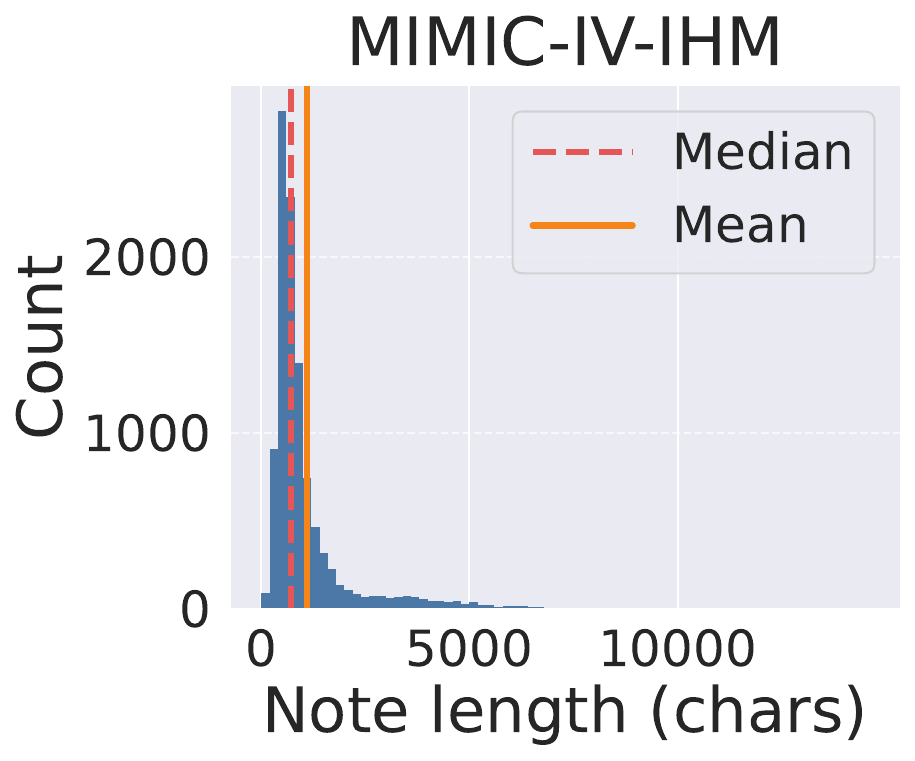}%
\hfill
\includegraphics[width=0.24\textwidth]{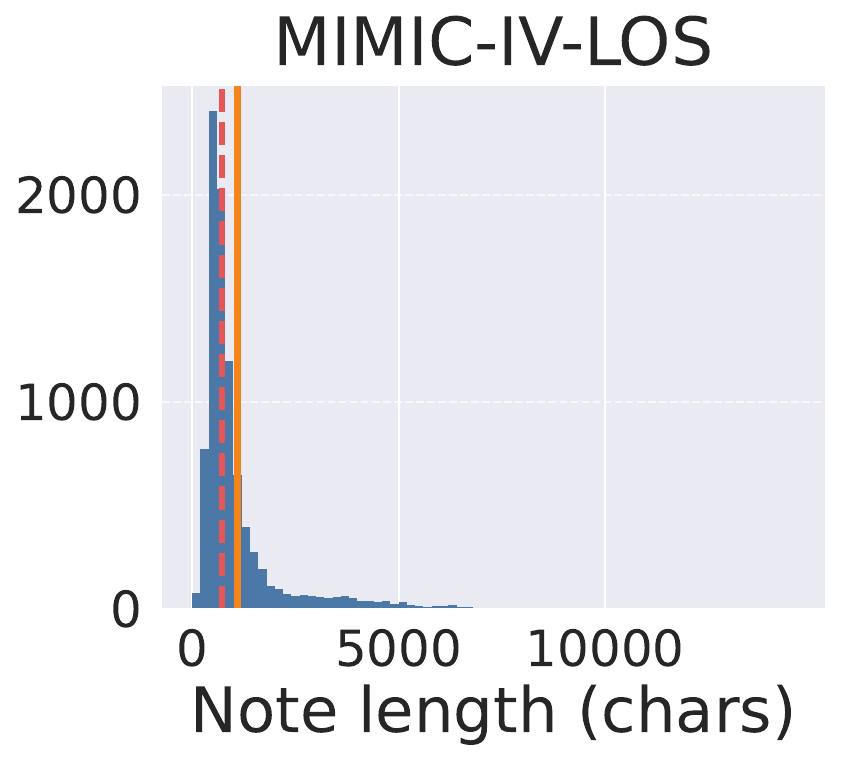}%
\hfill
\includegraphics[width=0.24\textwidth]{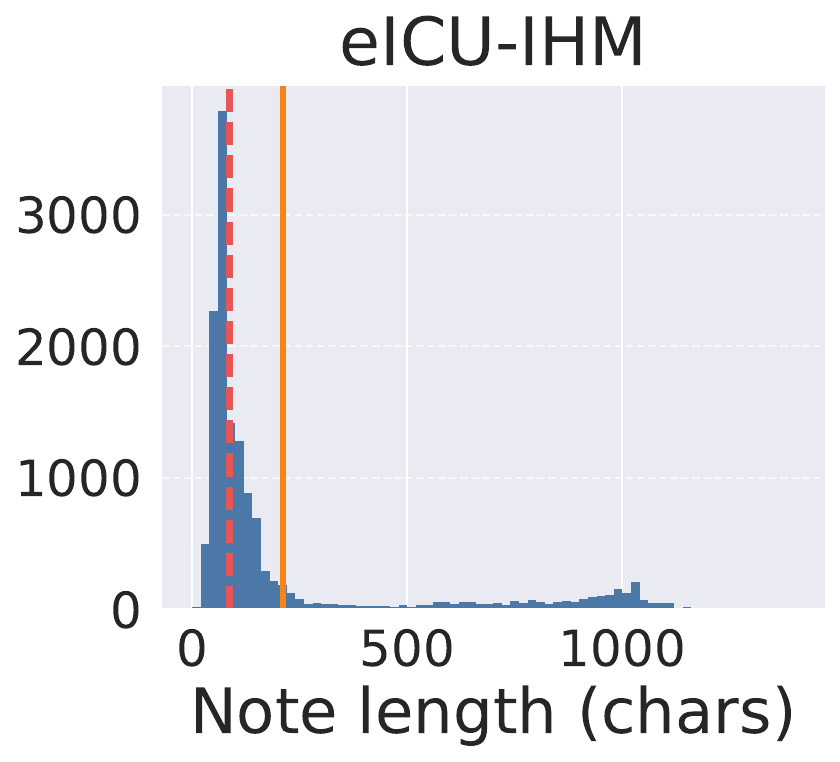}%
\hfill
\includegraphics[width=0.24\textwidth]{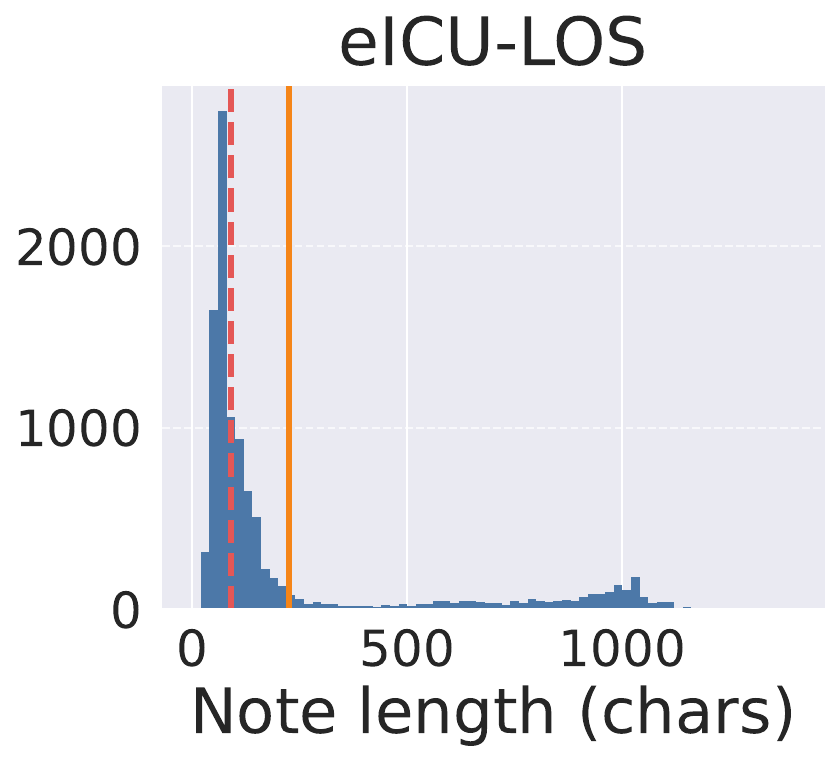}
\caption{Distribution of note character lengths across all four datasets. Dashed red and solid orange vertical lines indicate the median and mean, respectively.}
\label{fig:note_len_dist}
\end{figure}

We present the basic statistics of the $4$ datasets in Table~\ref{tab:dataset_stats}, including the number of MITS samples, positive rate, number of lab measurements per MITS, and number of clinical notes per MITS. We further report the number of lab measurements per sample for each channel in Table~\ref{tab:lab_channel_stats}. We also report the character count per note for each dataset to highlight differences in note properties between MIMIC-IV and eICU. The note length distributions are shown in Fig.~\ref{fig:note_len_dist}. Note lengths in both MIMIC-IV and eICU exhibit a long-tailed distribution. Notes in MIMIC-IV-IHM and MIMIC-IV-LOS can reach up to $5{,}000$ characters, whereas those in eICU-IHM and eICU-LOS reach only about $1{,}000$ characters, reflecting the richer free-text content of MIMIC-IV compared to eICU.

\section{Model and Hyper-Parameter Configurations}\label{app:model_config}

We present the model and hyper-parameter configurations of each method below. Hyper-parameters are selected based on validation AU-ROC.

\paragraph{MILM.} We fine-tune with QLoRA~\cite{dettmers2023qlora} (rank $r=16$, $\alpha=16$, dropout $0.05$) on the query, key, value, and output projection matrices. We use AdamW~\cite{loshchilov2018decoupled} with a learning rate of $10^{-4}$, cosine schedule, warmup ratio $0.05$, and effective batch size $8$. Stage~1 (value-redaction fine-tuning) trains for $8$ epochs across all datasets. The number of Stage~2 epochs is dataset-dependent, as shown below:

\medskip
\begin{center}
\scalebox{0.85}{
\begin{tabular}{lcccc}
\toprule
 & MIMIC-IV-IHM & MIMIC-IV-LOS & eICU-IHM & eICU-LOS \\
\midrule
\# Stage 2 Epochs & 5 & 6 & 8 & 10 \\
\bottomrule
\end{tabular}
}
\end{center}
\medskip

\paragraph{GRU-D.} GRU-D~\cite{che2018recurrent} extends the standard Gated Recurrent Unit with two trainable temporal decay mechanisms for irregularly sampled time series with missing values. Specifically, an input decay $\gamma_x$ decays each unobserved feature toward its empirical mean as a function of the elapsed time since its last observation, while a hidden state decay $\gamma_h$ similarly decays the GRU hidden state. The binary missingness mask is additionally fed into the GRU gates as an informative signal.

We use a hidden size of $32$ followed by an MLP with a single hidden layer of size $64$. Dropout and recurrent dropout are both set to $0.1$. The model is trained with the Adam optimizer \cite{kingma2015adam} at a learning rate of $10^{-3}$, with $L_2$ weight decay of $10^{-4}$ on the MLP hidden layer weights and gradient clipping at norm $1.0$. The batch size is $32$. The number of training epochs is dataset-dependent, as shown below:

\medskip
\begin{center}
\scalebox{0.85}{
\begin{tabular}{lcccc}
\toprule
 & MIMIC-IV-IHM & MIMIC-IV-LOS & eICU-IHM & eICU-LOS \\
\midrule
\# Epochs & 200 & 200 & 150 & 150 \\
\bottomrule
\end{tabular}
}    
\end{center}
\medskip

\paragraph{mTAND.} mTAND~\cite{shukla2021multitime} models irregularly sampled time series by learning a continuous-time embedding of observation timestamps and using a multi-head attention mechanism to interpolate observed values onto a fixed set of reference time points, producing a fixed-length representation without requiring imputation or temporal discretization. For classification, we use the encoder-only variant (mTAND-Enc), where the mTAND module maps irregular observations to $K$ reference-point embeddings that are then aggregated by a GRU, followed by a two-layer residual projection head.

We use an attention embedding dimension of $64$, a time embedding dimension of $32$, $8$ attention heads, $K=24$ reference points (matching the 24-hour observation window), and a GRU hidden size of $128$. Dropout is set to $0.3$. The model is trained with the Adam optimizer with gradient clipping at norm $1.0$ and a batch size of $32$ for $250$ epochs across all datasets. The learning rate is dataset-dependent, as shown below:

\medskip
\begin{center}
\scalebox{0.85}{
\begin{tabular}{lcccc}
\toprule
 & MIMIC-IV-IHM & MIMIC-IV-LOS & eICU-IHM & eICU-LOS \\
\midrule
Learning Rate & $2\times10^{-4}$ & $1\times10^{-4}$ & $2\times10^{-4}$ & $1\times10^{-4}$ \\
\bottomrule
\end{tabular}
}
\end{center}
\medskip

\paragraph{UTDE.}
UTDE~\cite{zhang2023improving} is a multimodal framework for irregular EHR time series that unifies two complementary embedding strategies via a learned gate: an imputation-based embedding (forward-fill followed by a 1-D convolution) and an mTAND-based interpolation embedding, yielding a unified time series representation. Clinical notes are cast as multivariate irregular time series and encoded through a separate mTAND module. The two modalities are then fused through $J$ identical layers of interleaved self-attention and cross-attention. We use the \inlinecode{self\_cross} fusion variant with Adam optimizer, a learning rate of $2\times10^{-4}$, batch size 32, gradient clipping at 1.0, dropout 0.3, $d_{\text{time}}=32$ time embedding dimensions, 8 attention heads, 2 self-attention layers, and 2 cross-attention layers. Dataset-specific hyper-parameters are summarized below:

\medskip
\begin{center}
\label{tab:utde_hparams}
\scalebox{0.85}{
\begin{tabular}{@{}lcccc@{}}
\toprule
Hyper-Parameter & MIMIC-IV-IHM & MIMIC-IV-LOS & eICU-IHM & eICU-LOS \\
\midrule
Embedding Dim. ($d$) & 64 & 64 & 32 & 64 \\
Max. \# Notes Used   & 5  & 5  & 9  & 9  \\
\# Epochs            & 150 & 150 & 250 & 250 \\
\bottomrule
\end{tabular}
}
\end{center}
\medskip

\paragraph{FuseMoE.}
FuseMoE~\citep{han2024fusemoe} shares the per-modality encoding backbone with UTDE, using mTAND for irregularly sampled time series and a projection layer for clinical note embeddings, but replaces the cross-attention fusion with a sparse Mixture-of-Experts (MoE) fusion layer. In each layer, a joint router receives the concatenated embeddings of all modalities and dispatches each sample to the top-$K$ of $E$ expert MLPs. We use Adam with a learning rate of $2\times10^{-4}$, batch size 32, gradient clipping at 1.0, dropout 0.3, $d=64$ embedding dimensions, $d_{\text{time}}=32$ time embedding dimensions, 8 attention heads, 2 self-attention layers, 2 cross-encoder layers, 3 experts, top-$K{=}2$ routing, a joint router, MoE hidden size 512, and a load-balance auxiliary loss with coefficient 0.01. Dataset-specific hyper-parameters are summarized below:

\medskip
\begin{center}
\label{tab:fusemoe_hparams}
\scalebox{0.85}{
\begin{tabular}{@{}lcccc@{}}
\toprule
Hyper-Parameter & MIMIC-IV-IHM & MIMIC-IV-LOS & eICU-IHM & eICU-LOS \\
\midrule
Max. \# Notes Used & 5  & 5  & 9  & 9  \\
\# Epochs     & 100 & 100 & 150 & 150 \\
\bottomrule
\end{tabular}
}
\end{center}
\medskip

\paragraph{VITAL-stats \& VITAL-LLM.}
VITAL~\citep{kwon2025mind} is a variable-aware framework that differentiates between two types of clinical variables: frequently measured vital signs processed through LLM reprogramming, and infrequently measured lab tests embedded via per-channel summary statistics or a learnable \texttt{[Not measured]} token.
Since the numerical channels of our datasets contain only laboratory measurements, we evaluate both branches separately: VITAL-stats applies the lab branch, representing each channel by a four-dimensional summary vector (mean, median, min, max) followed by feature mixing. VITAL-LLM applies the vital sign reprogramming branch, which requires regular-grid input and is therefore applied to an hourly-discretized version of the lab channels. For fairness of comparison, VITAL-LLM uses the same backbone LLM as MILM (Qwen3-4B-Instruct-2507).

Both variants are trained with Adam, learning rate $1\times10^{-3}$, dropout 0.1, and gradient clipping at 1.0.
VITAL-stats uses batch size 128 and $d_{\text{ff}}=32$ uniformly.
VITAL-LLM uses batch size 8 with 16 gradient accumulation steps (effective batch size 128), 1{,}000 text prototypes, 6 reprogramming attention heads, and bfloat16 precision.
Dataset-specific hyper-parameters are summarized in Tables~\ref{tab:vital_stats_hparams} and~\ref{tab:vital_llm_hparams}.

\begin{table}[h]
\centering
\caption{Dataset-specific hyper-parameters for VITAL-stats.}
\label{tab:vital_stats_hparams}
\scalebox{0.85}{
\begin{tabular}{@{}lcccc@{}}
\toprule
Hyper-Parameter & MIMIC-IV-IHM & MIMIC-IV-LOS & eICU-IHM & eICU-LOS \\
\midrule
\# Epochs & 50 & 100 & 50 & 100 \\
\bottomrule
\end{tabular}
}
\end{table}

\begin{table}[h]
\centering
\caption{Dataset-specific hyper-parameters for VITAL-LLM.}
\label{tab:vital_llm_hparams}
\scalebox{0.85}{
\begin{tabular}{@{}lcccc@{}}
\toprule
Hyper-Parameter & MIMIC-IV-IHM & MIMIC-IV-LOS & eICU-IHM & eICU-LOS \\
\midrule
$d_{\text{ff}}$ & 32 & 64 & 32 & 64 \\
\bottomrule
\end{tabular}
}
\end{table}

\paragraph{ISTS-PLM.}
ISTS-PLM~\cite{zhang2025unleashing} adopts a series-based representation of irregularly sampled time series, processing each variable's observations as a univariate sequence through two successive pretrained language model (PLM) stages. The first stage is a time-aware PLM that replaces the absolute positional embeddings with learnable continuous-time embeddings, enabling the model to reason about intra-series temporal dynamics. The second stage is a variable-aware PLM that replaces positional embeddings with learnable variable embeddings, capturing inter-series correlations across asynchronous channels. We use GPT-2~\cite{radford2019language} as the time-aware PLM and BERT~\cite{devlin2019bert} as the variable-aware PLM, following the original paper. Applying Qwen3 is not feasible because Qwen3 uses rotary position embeddings (RoPE)~\cite{su2024roformer}, which are fused into the attention computation and cannot be straightforwardly replaced by continuous-time embeddings. We use the first 6 layers of each PLM ($d_\text{model}=768$), follow the original paper to freeze all PLM parameters except layer normalization, use dropout 0.1, Adam optimizer, batch size 6, and gradient clipping at $1.0$ for $40$ epochs. Dataset-specific hyper-parameters are summarized below:
\medskip
\begin{center}
\label{tab:ists_plm_hparams}
\scalebox{0.85}{
\begin{tabular}{@{}lcccc@{}}
\toprule
Hyper-Parameter & MIMIC-IV-IHM & MIMIC-IV-LOS & eICU-IHM & eICU-LOS \\
\midrule
Learning rate & $2\times10^{-3}$ & $1\times10^{-3}$ & $1\times10^{-3}$ & $5\times10^{-4}$ \\
\bottomrule
\end{tabular}
}
\end{center}
\medskip

\begin{table}[t]
\centering
\caption{Total wall-clock training time (minutes) for baselines, run on an NVIDIA RTX 3090.}
\label{tab:training_time_baselines}
\scalebox{0.85}{
\begin{tabular}{lcccc}
\toprule
 & \textbf{MIMIC-IV-IHM} & \textbf{MIMIC-IV-LOS} & \textbf{eICU-IHM} & \textbf{eICU-LOS} \\
\midrule
GRU-D                        & 15.00  & 12.88  & 8.02   & 4.43   \\
\quad {\footnotesize + Note} & 17.12  & 14.70  & 7.22   & 5.43   \\
mTAND                        & 3.82   & 3.22   & 2.98   & 2.23   \\
\quad {\footnotesize + Note} & 6.05   & 5.33   & 5.90   & 4.25   \\
UTDE                         & 6.20   & 5.55   & 11.87  & 11.98  \\
FuseMoE                      & 5.03   & 5.13   & 11.86  & 11.82  \\
VITAL-stats                  & 0.50   & 0.85   & 0.35   & 0.50   \\
\quad {\footnotesize + Note} & 0.78   & 1.40   & 0.72   & 1.00   \\
VITAL-LLM                    & 428.78 & 372.51 & 352.32 & 121.15 \\
\quad {\footnotesize + Note} & 698.17 & 359.30 & 465.97 & 167.53 \\
ISTS-PLM                     & 8.67   & 23.06  & 8.42   & 9.62   \\
\quad {\footnotesize + Note} & 25.70  & 21.63  & 18.97  & 18.60  \\
\bottomrule
\end{tabular}
}
\end{table}

\paragraph{Timestamp-to-Text Fusion (TTF) Module.}
The Timestamp-to-Text Fusion (TTF) module, introduced in Time-IMM~\citep{chang2025timeimm}, addresses the challenge of integrating asynchronously timestamped clinical notes with numerical time series by producing a temporally-aware and fixed-size text representation. We adopt the T2V-XAttn variant. Each note embedding is concatenated with a Time2Vec encoding~\cite{kazemi2019time2vec} of its timestamp, projected to a common dimension, and then attended to by a single learned query vector via single-head cross-attention, yielding a pooled text representation that is conditioned on both the semantic content and the temporal position of each note. This representation is concatenated to the representation of numerical observations before the classification head. We apply TTF to augment GRU-D, mTAND, VITAL-stats, VITAL-LLM, and ISTS-PLM, producing the corresponding ``+~Note'' baseline variants reported in the main table. All TTF encoders use 768-dimensional BioBERT~\cite{lee2020biobert} note embeddings. We use the 6 most recent notes per patient on MIMIC-IV and the 10 most recent notes per patient on eICU.

\begin{table}[t]
\centering
\caption{Total wall-clock training time (minutes) for MILM, run on an NVIDIA H200.}
\label{tab:training_time_milm}
\scalebox{0.85}{
\begin{tabular}{lcccc}
\toprule
 & \textbf{MIMIC-IV-IHM} & \textbf{MIMIC-IV-LOS} & \textbf{eICU-IHM} & \textbf{eICU-LOS} \\
\midrule
MILM (Stage 1) & 254.38 & 207.07 & 306.73 & 206.18 \\
MILM (Stage 2) & 162.00 & 168.83 & 311.45 & 212.18 \\
\bottomrule
\end{tabular}
}
\end{table}

\begin{table}[t]
\centering
\caption{Peak GPU memory usage (GB).}
\label{tab:peak_gpu_memory}
\scalebox{0.85}{
\begin{tabular}{lcccc}
\toprule
 & \textbf{MIMIC-IV-IHM} & \textbf{MIMIC-IV-LOS} & \textbf{eICU-IHM} & \textbf{eICU-LOS} \\
\midrule
GRU-D                        & 0.03  & 0.03  & 0.02  & 0.02  \\
\quad {\footnotesize + Note} & 0.11  & 0.11  & 0.10  & 0.10  \\
mTAND                        & 0.38  & 0.58  & 0.20  & 0.12  \\
\quad {\footnotesize + Note} & 0.34  & 0.63  & 0.31  & 0.23  \\
UTDE                         & 0.59  & 0.62  & 0.95  & 0.79  \\
FuseMoE                      & 0.85  & 0.85  & 1.17  & 1.19  \\
VITAL-stats                  & 0.03  & 0.03  & 0.03  & 0.03  \\
\quad {\footnotesize + Note} & 0.12  & 0.12  & 0.12  & 0.12  \\
VITAL-LLM                    & 11.16 & 11.18 & 11.22 & 11.24 \\
\quad {\footnotesize + Note} & 11.27 & 11.29 & 11.35 & 11.37 \\
ISTS-PLM                     & 4.87  & 2.86  & 3.13  & 2.81  \\
\quad {\footnotesize + Note} & 4.89  & 3.75  & 3.31  & 2.46  \\
\midrule
MILM (Stage 1)               & 48.96 & 44.99 & 26.48 & 34.68 \\
MILM (Stage 2)               & 64.20 & 71.20 & 39.93 & 52.43 \\
\bottomrule
\end{tabular}
}
\end{table}

\section{Compute Resources}\label{app:compute_resource}

All baseline experiments are run on an NVIDIA GeForce RTX 3090 (24~GB). MILM requires more GPU memory due to LLM fine-tuning, and is therefore run on an NVIDIA H200 GPU (96~GB) on a GH200 node. This overhead reflects the per-layer LoRA fine-tuning that genuinely adapts the LLM's weights to domain-specific temporal knowledge and gives MILM the capacity to predict under either setting: from sampling patterns alone or from both patterns and values. In contrast, the LLM-based irregular time series baselines keep the LLM weights frozen by design and train surrounding modules. We report total wall-clock training time per run in Tables~\ref{tab:training_time_baselines} and~\ref{tab:training_time_milm}, and peak GPU memory usage in Table~\ref{tab:peak_gpu_memory}.

\section{Value Pending Evaluation Details}\label{app:value_pending_eval}

\subsection{Dataset Statistics and Analyses}

\begin{table}[t]
\centering
\caption{Average number (left) and rate (right) of pending observations per ICU stay.}
\label{tab:pending_summary}
\begin{minipage}[t]{0.42\linewidth}
\centering
\scalebox{0.85}{
\begin{tabular}{lcc}
\toprule
& \textbf{MIMIC-IV-IHM} & \textbf{MIMIC-IV-LOS} \\
\midrule
Lab            & 2.74 & 2.71 \\
Note           & 0.43 & 0.45 \\
\bottomrule
\end{tabular}
}
\end{minipage}%
\hspace{3em}%
\begin{minipage}[t]{0.42\linewidth}
\centering
\scalebox{0.85}{
\begin{tabular}{lcc}
\toprule
& \textbf{MIMIC-IV-IHM} & \textbf{MIMIC-IV-LOS} \\
\midrule
Lab            &  3.33\% &  3.30\% \\
Note & 13.53\% & 14.02\% \\
\bottomrule
\end{tabular}
}
\end{minipage}
\end{table}

\begin{table}[h!]
\centering
\caption{Average and median storetime $-$ charttime delay (hours) per channel across all ICU stays.}
\label{tab:delay_per_channel}
\scalebox{0.85}{
\begin{tabular}{lcccc}
\toprule
& \multicolumn{2}{c}{\textbf{MIMIC-IV-IHM}}
& \multicolumn{2}{c}{\textbf{MIMIC-IV-LOS}} \\
\cmidrule(lr){2-3}
\cmidrule(lr){4-5}
\textbf{Channel} & Avg. & Median & Avg. & Median \\
\midrule
Anion Gap          & 1.26 & 1.13 & 1.25 & 1.12 \\
Bicarbonate        & 1.24 & 1.12 & 1.23 & 1.10 \\
Calcium, Total     & 1.20 & 1.08 & 1.20 & 1.08 \\
Chloride           & 1.22 & 1.10 & 1.22 & 1.08 \\
Creatinine         & 1.22 & 1.08 & 1.21 & 1.08 \\
Glucose            & 0.09 & 0.05 & 0.09 & 0.05 \\
Hematocrit         & 0.61 & 0.52 & 0.61 & 0.52 \\
Hemoglobin         & 0.10 & 0.07 & 0.10 & 0.07 \\
MCH                & 0.62 & 0.53 & 0.61 & 0.52 \\
MCHC               & 0.62 & 0.53 & 0.61 & 0.52 \\
MCV                & 0.62 & 0.53 & 0.61 & 0.52 \\
Magnesium          & 1.21 & 1.08 & 1.20 & 1.08 \\
Neutrophils        & 1.27 & 0.87 & 1.26 & 0.87 \\
Phosphate          & 1.21 & 1.10 & 1.21 & 1.08 \\
Platelet Count     & 0.66 & 0.55 & 0.66 & 0.53 \\
RDW                & 0.62 & 0.53 & 0.61 & 0.52 \\
Red Blood Cells    & 0.62 & 0.53 & 0.61 & 0.52 \\
Sodium             & 1.23 & 1.10 & 1.22 & 1.10 \\
Urea Nitrogen      & 1.19 & 1.08 & 1.19 & 1.08 \\
Vancomycin         & 2.31 & 1.58 & 2.31 & 1.59 \\
White Blood Cells  & 0.64 & 0.53 & 0.64 & 0.53 \\
Radiology Note     & 5.73 & 3.17 & 5.74 & 3.23 \\
\bottomrule
\end{tabular}
}
\end{table}

\begin{figure}[t]
\centering
\includegraphics[width=0.26\textwidth]{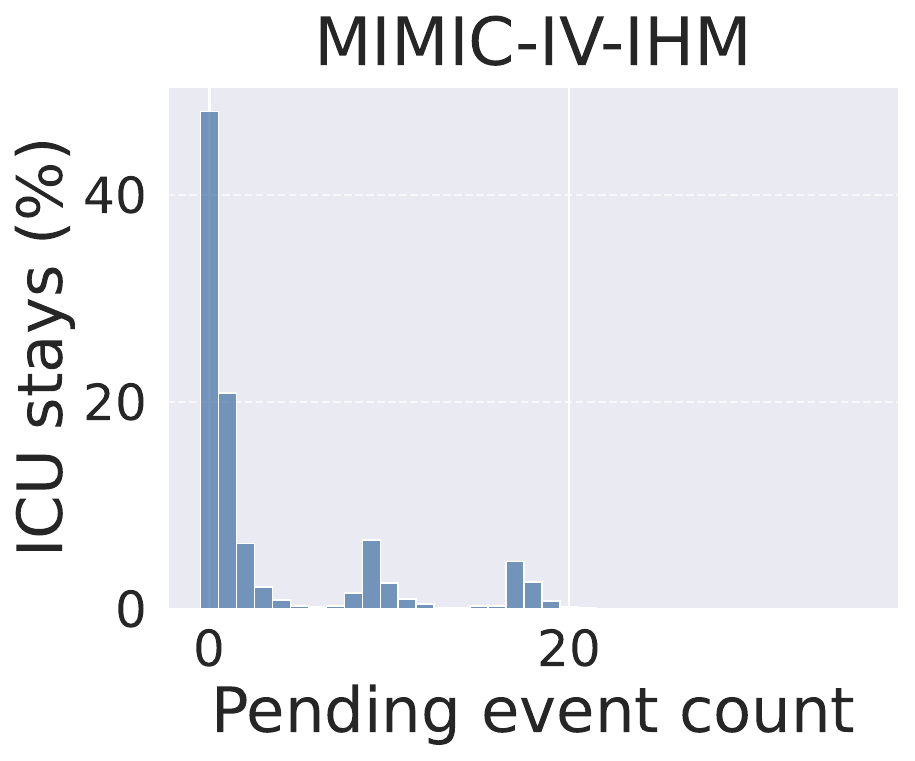}%
\hfill
\includegraphics[width=0.24\textwidth]{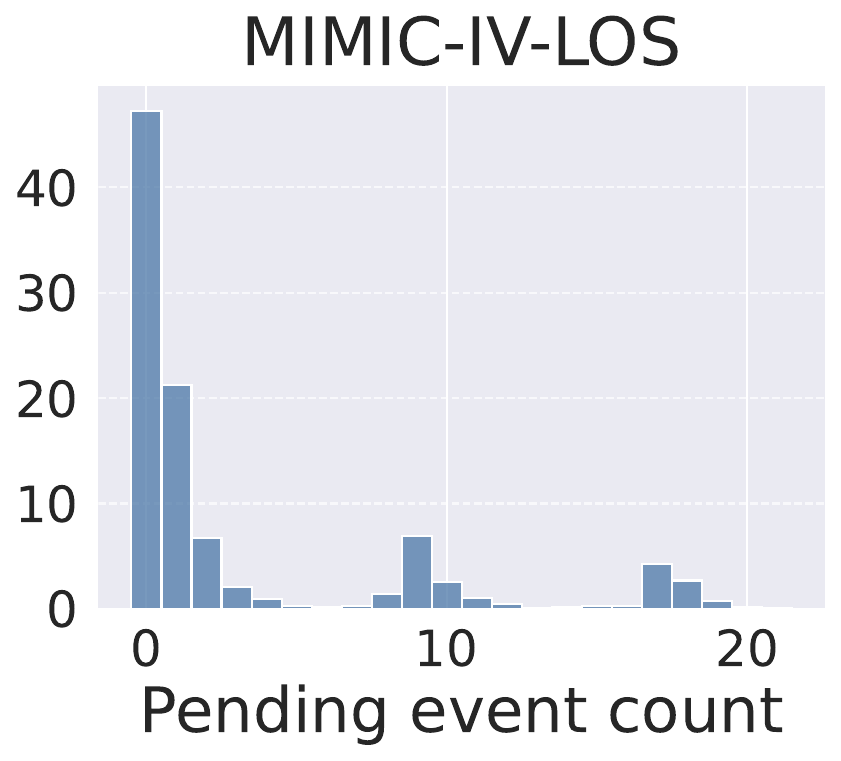}%
\hfill
\includegraphics[width=0.24\textwidth]{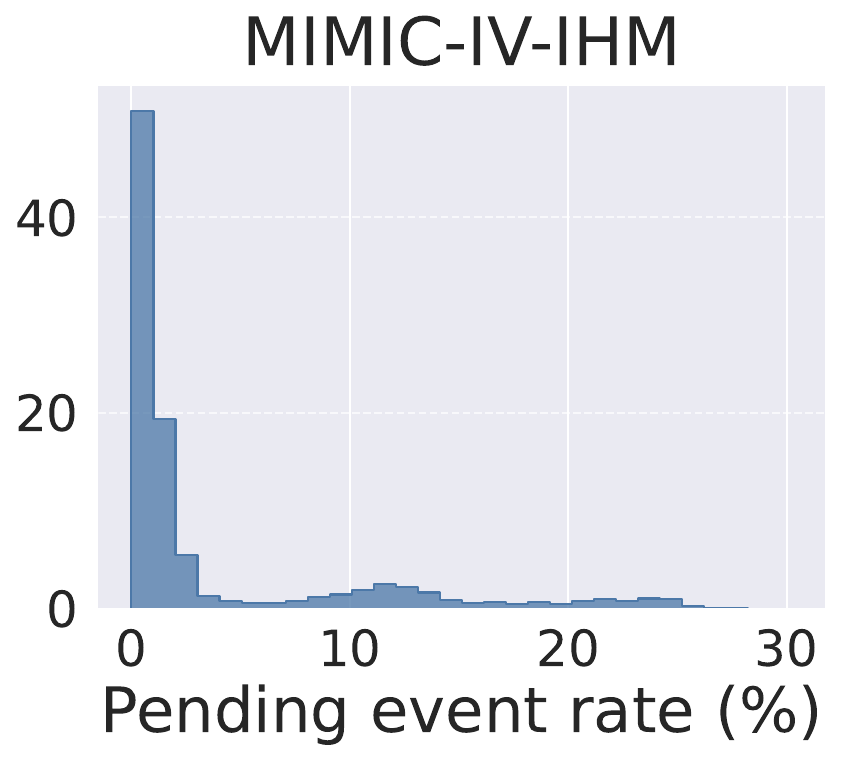}%
\hfill
\includegraphics[width=0.24\textwidth]{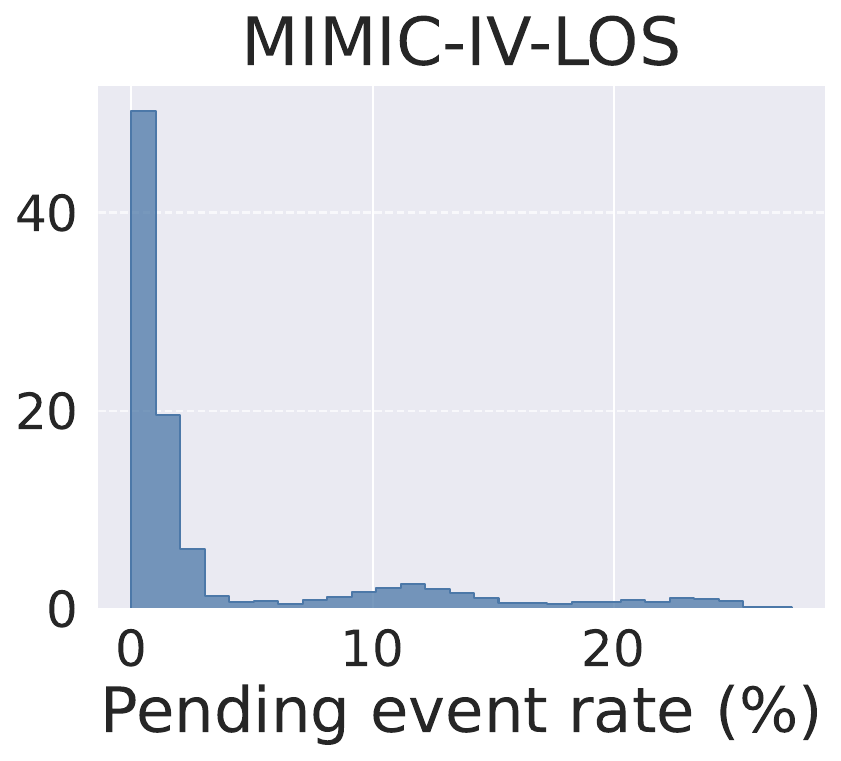}
\caption{Distribution of pending event count (left two) and pending event rate (right two) per ICU stay at the 24-hour prediction time.}
\label{fig:pending_dist}
\end{figure}

\begin{figure}[t]
\centering
\includegraphics[width=0.26\textwidth]{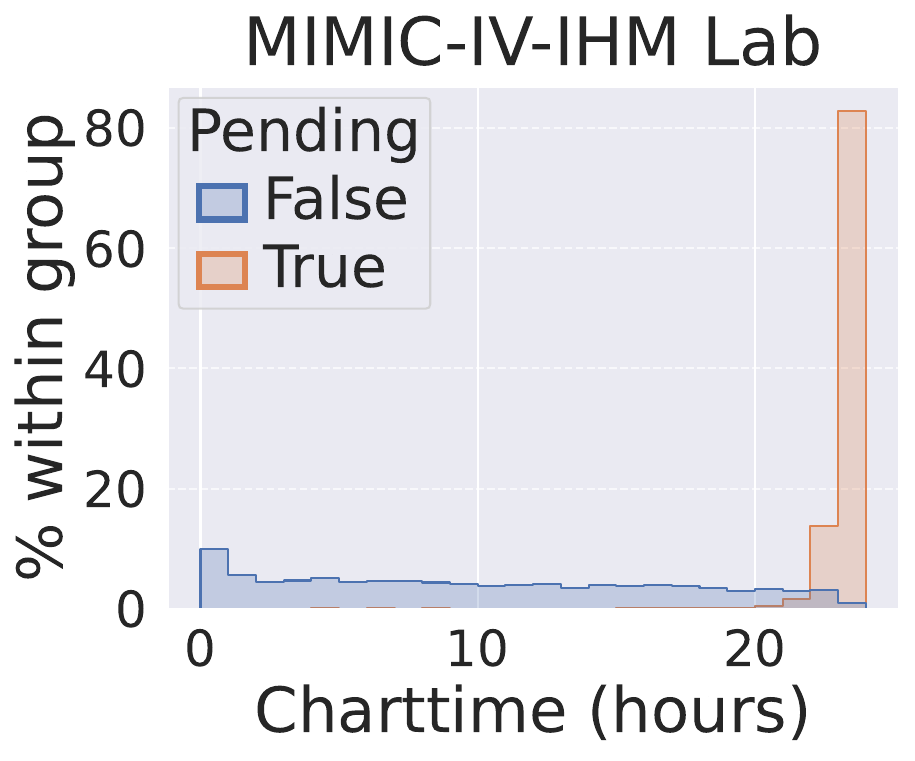}%
\hfill
\includegraphics[width=0.24\textwidth]{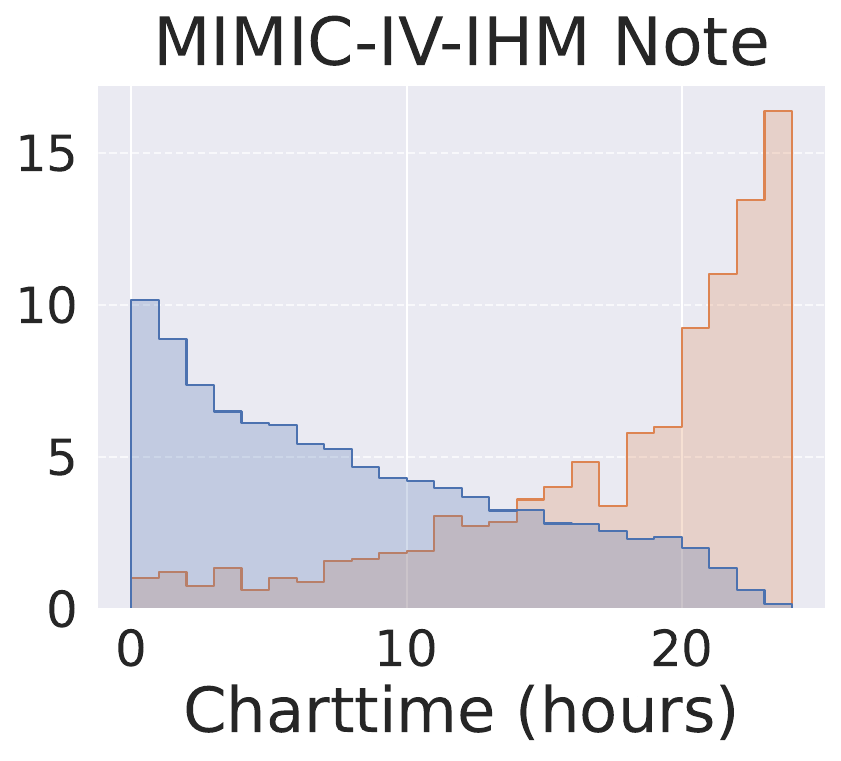}%
\hfill
\includegraphics[width=0.24\textwidth]{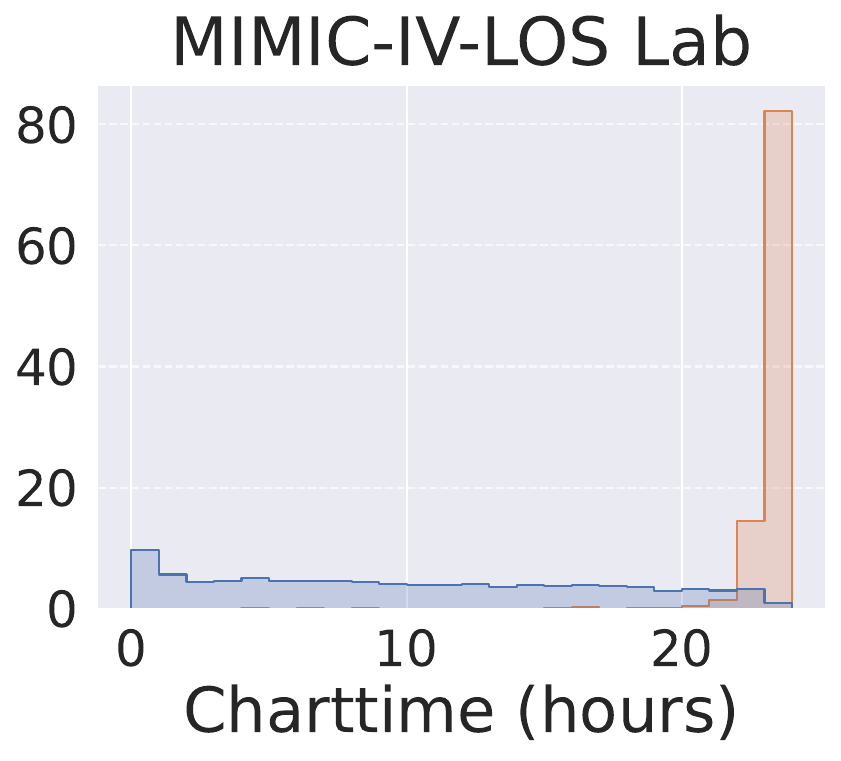}%
\hfill
\includegraphics[width=0.24\textwidth]{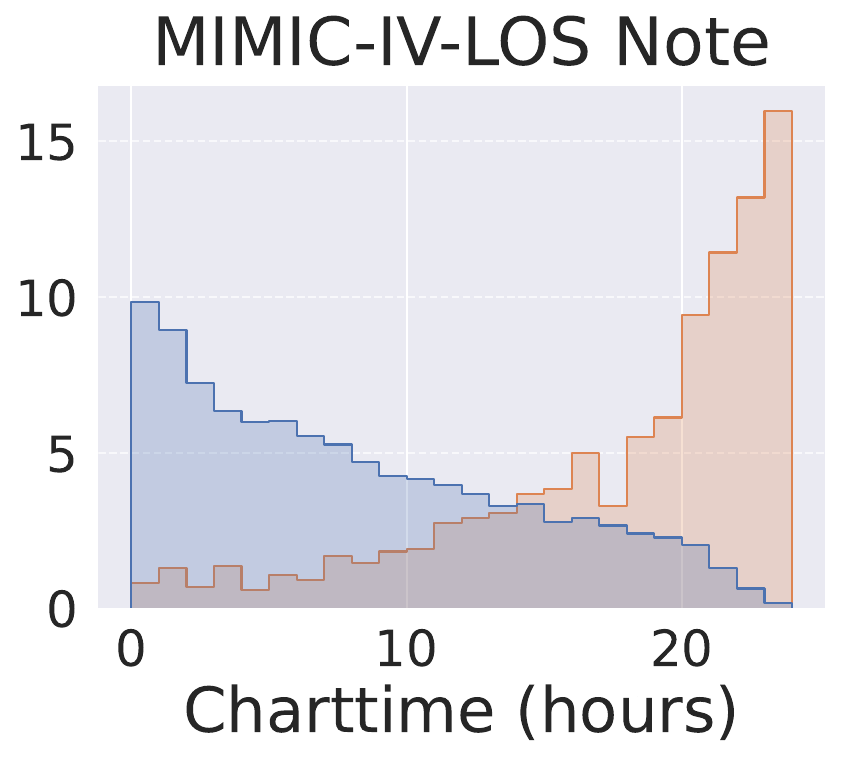}
\caption{Distribution of \inlinecode{charttime} for value-pending and non-pending observations.}
\label{fig:pending_charttime_dist}
\end{figure}

For the value pending evaluation in Section~\ref{subsec:value_pending}, an observation's value is considered pending if its \inlinecode{storetime}, the timestamp at which the result was available in the database, exceeds the 24-hour prediction time. When \inlinecode{storetime} is absent, we conservatively fall back to \inlinecode{charttime}, treating the observation as available on time. The same rule is applied to radiology notes.

Table~\ref{tab:pending_summary} reports the average number and rate of pending observations per ICU stay. On average, each stay has approximately 2.74 (IHM) and 2.71 (LOS) pending lab measurements, representing about 3.3\% of all lab observations. Notes are pending at a higher rate: 13.5\% (IHM) and 14.0\% (LOS). Table~\ref{tab:delay_per_channel} shows the average and median storetime $-$ charttime delay per channel. Lab delays cluster around 0.6 to 1.3 hours depending on the panel, while note delays are markedly longer (median $\approx$ 3.2 hours), consistent with the higher note pending rate.

Figure~\ref{fig:pending_dist} shows how these counts and rates vary across individual ICU stays. Both are right-skewed: the majority of stays have few or no pending observations, but a nontrivial tail of stays can have 10 to 30\% of their observations pending.

Figure~\ref{fig:pending_charttime_dist} shows $p(\text{\inlinecode{charttime}} \mid \text{value pending})$ and $p(\text{\inlinecode{charttime}} \mid \text{value not pending})$ separately for labs and notes. The pending distributions are visibly right-shifted relative to the non-pending distributions, confirming that value-pending observations typically fall near the end of the 24-hour observation window, where recent measurements tend to be most informative for outcome prediction.

\subsection{Full Results}
\begin{table}[t]
\centering
\caption{Full value pending evaluation results. Best results are \textbf{boldfaced} and second-best results are \underline{underlined}.}
\label{tab:value_pending_full}
\scalebox{0.9}{
\begin{tabular}{llcccc}
\toprule
& & \multicolumn{2}{c}{\textbf{MIMIC-IV-IHM}} & \multicolumn{2}{c}{\textbf{MIMIC-IV-LOS}} \\
\cmidrule(lr){3-4} \cmidrule(lr){5-6}
\textbf{Countermeasure} & \textbf{Method} & AU-ROC & AP & AU-ROC & AP \\
\midrule
Drop Observation
& GRU-D                        & \std{75.07}{1.03} & \std{52.61}{1.07} & \std{69.14}{1.20} & \std{46.37}{1.43} \\
& \quad {\footnotesize + Note} & \std{73.60}{1.79} & \std{50.21}{4.12} & \std{70.12}{2.56} & \std{47.51}{2.46} \\
& mTAND                        & \std{76.56}{1.14} & \std{54.63}{1.70} & \std{71.19}{0.83} & \std{49.83}{1.41} \\
& \quad {\footnotesize + Note} & \std{77.46}{1.55} & \std{56.54}{1.05} & \std{73.02}{1.02} & \std{52.86}{2.83} \\
& UTDE                         & \std{74.22}{2.93} & \std{52.54}{3.02} & \std{73.89}{3.17} & \std{52.80}{3.16} \\
& FuseMoE                      & \std{75.49}{1.47} & \std{55.00}{0.92} & \std{74.56}{0.46} & \std{55.10}{1.75} \\
& VITAL-stats                  & \std{76.44}{0.84} & \std{53.14}{2.13} & \std{73.70}{0.95} & \std{52.58}{2.72} \\
& \quad {\footnotesize + Note} & \std{77.10}{0.63} & \std{52.86}{1.99} & \std{74.24}{1.25} & \std{54.47}{4.03} \\
& VITAL-LLM                    & \std{67.96}{1.61} & \std{40.66}{2.00} & \std{68.77}{1.59} & \std{47.70}{3.36} \\
& \quad {\footnotesize + Note} & \std{67.98}{2.77} & \std{40.42}{3.07} & \std{68.80}{1.26} & \std{48.37}{1.83} \\
& ISTS-PLM                     & \second{79.95}{0.59} & \std{59.31}{0.77} & \std{73.93}{0.73} & \std{54.27}{1.59} \\
& \quad {\footnotesize + Note} & \std{79.52}{0.53} & \std{58.22}{0.64} & \std{73.63}{0.51} & \std{52.46}{0.92} \\
& Qwen3-4B                     & \std{68.33}{0.00} & \std{44.06}{0.00} & \std{67.39}{0.00} & \std{44.37}{0.00} \\
& MILM-Direct                  & \std{79.04}{0.91} & \std{59.23}{1.63} & \std{76.86}{0.52} & \std{56.44}{1.61} \\
& MILM-2S                      & \std{79.49}{0.76} & \second{60.02}{1.57} & \best{77.64}{0.33} & \best{57.17}{1.78} \\
\midrule
Show Presence
& Qwen3-4B                     & \std{67.40}{0.00} & \std{39.80}{0.00} & \std{66.54}{0.00} & \std{43.50}{0.00} \\
& MILM-Direct                  & \std{79.21}{0.65} & \std{58.14}{1.37} & \std{76.30}{0.64} & \std{55.01}{2.07} \\
& MILM-2S                      & \best{80.34}{0.64} & \best{61.23}{1.62} & \second{77.08}{0.38} & \second{56.46}{1.50} \\
\bottomrule
\end{tabular}
}
\end{table}

We present the full value pending evaluation results in Table~\ref{tab:value_pending_full}, extending Table~\ref{tab:value_pending} to include baselines with the drop observation countermeasure. Note that it is not straightforward to make the baselines show the presence of value-pending observations without showing the values, as these models cannot process \inlinecode{NaN} values. Qwen3-4B and MILM are able to adopt the show presence countermeasure due to the flexibility of text-based representations.

Value pending evaluation has a larger impact on natively multimodal methods such as UTDE, Qwen3-4B, and MILM since the note channel has a higher pending rate than lab channels (Table ~\ref{tab:pending_summary}). However, the two-stage training of MILM-2S effectively mitigates this challenge. By training the model to predict solely from sampling patterns, Stage 1 develops two complementary capabilities. First, by learning to extract predictive signals without value tokens, MILM-2S makes better use of the remaining observations under drop observation when some note content is missing. Second, by learning to treat time-channel patterns as signals in their own right, MILM-2S makes better use of the additional sampling information conveyed under show presence. Both capabilities translate directly to stronger performance than MILM-Direct under value pending evaluation. Despite operating under a more significant train-test discrepancy than unimodal baselines, MILM-2S with show presence and MILM-2S with drop observation remain the best-performing methods on MIMIC-IV-IHM and MIMIC-IV-LOS, respectively.

\section{MedGemma vs. Qwen}

Due to the domain of interest, it is natural for us to consider using large language models pretrained for the biomedical and healthcare domains, such as
MedGemma~\cite{sellergren2025medgemma}. MedGemma is a collection of medical vision-language foundation models built on the Gemma~3 architecture, designed to interpret and reason about medical images and text across modalities including radiology, dermatology, histopathology, and ophthalmology. In our preliminary experiments, we compare the zero-shot performance of the $4$B instruction-tuned variants of both MedGemma and Qwen3. We report the results below:

\begin{center}
\scalebox{0.85}{
\begin{tabular}{lcccccc}
\toprule
& \multicolumn{3}{c}{MIMIC-IV-IHM} & \multicolumn{3}{c}{MIMIC-IV-LOS} \\
\cmidrule(lr){2-4} \cmidrule(lr){5-7}
Model & Unparseable (\%) & AU-ROC & AP & Unparseable (\%) & AU-ROC & AP \\
\midrule
MedGemma-4B & 62.70 & 57.81 & 36.27 & 89.66 & 52.66 & 33.23 \\
Qwen3-4B    & \textbf{0.00}  & \textbf{69.82} & \textbf{48.36} & \textbf{0.00}  & \textbf{67.98} & \textbf{44.82} \\
\bottomrule
\end{tabular}
}
\end{center}

\begin{center}
\scalebox{0.85}{
\begin{tabular}{lcccccc}
\toprule
& \multicolumn{3}{c}{eICU-IHM} & \multicolumn{3}{c}{eICU-LOS} \\
\cmidrule(lr){2-4} \cmidrule(lr){5-7}
Model & Unparseable (\%) & AU-ROC & AP & Unparseable (\%) & AU-ROC & AP \\
\midrule
MedGemma-4B & 65.70 & 52.47 & 22.40 & 80.64 & 53.38 & 43.33 \\
Qwen3-4B    & \textbf{0.00}  & \textbf{64.53} & \textbf{29.43} & \textbf{0.00}  & \textbf{55.71} & \textbf{45.58} \\
\bottomrule
\end{tabular}
}
\end{center}

A large portion of the responses generated by MedGemma-4B are unparseable. For example, instead of generating \inlinecode{<answer> A </answer>}, it may generate \inlinecode{A}, violating the output format specified in the prompt. In contrast, the responses from Qwen3-4B are all parseable, demonstrating its stronger instruction-following capability and stability. We evaluate AU-ROC and AP by assigning a neutral score of 0.5 to unparseable responses to maintain a consistent evaluation cohort. Despite being pretrained on biomedical and healthcare domain data, MedGemma-4B underperforms Qwen3-4B on all 8 metric-dataset combinations. We suspect this is because MedGemma's domain-specific training is primarily oriented toward medical image understanding, with text supervision derived from relatively small medical QA datasets, whereas Qwen3 is pretrained on a substantially larger and more diverse corpus of 36 trillion tokens across 119 languages, which may better support the text representation inputs used in our task. Given the stronger instruction-following capability and better overall performance of Qwen3-4B, we select it as the base LLM for this work.

\section{Limitations}\label{app:limitations}

MILM provides a framework for using LLMs to make predictions from MITS with informative sampling. Although the method can be extended to image channels when the LLM includes a vision encoder and image content is serialized as image tokens, we do not consider image modalities in this work due to the substantially greater computational cost compared with processing numerical and text channels using text-only LLMs. We leave the extension to additional modalities for future work.

\section{Broader Impacts}\label{app:impact}
MILM is a research framework for applying LLMs to multimodal irregular time series classification in clinical settings. On the positive side, better usage of informative sampling patterns in ICU data could support the development of improved clinical prediction systems. On the negative side, the models in this work are trained and evaluated on US-based EHR datasets (MIMIC-IV and eICU), and may not generalize equally across hospital systems, countries, or patient populations. We emphasize that MILM is not validated for direct clinical deployment, and any downstream application would require rigorous prospective validation.

\section{Asset Licenses}\label{app:licenses}

\paragraph{Datasets.} MIMIC-IV~\cite{johnson2023mimic,PhysioNet-mimiciv-3.1}, MIMIC-IV-Note~\cite{PhysioNet-mimic-iv-note-2.2}, and the eICU Collaborative Research Database~\cite{pollard2018eicu,PhysioNet-eicu-crd-2.0} are available via PhysioNet under the PhysioNet Credentialed Health Data License 1.5.0. All datasets are used in compliance with their respective data use agreements.

\paragraph{Models.} Qwen3-4B-Instruct-2507~\cite{yang2025qwen3}, BioBERT~\cite{lee2020biobert}, and BERT~\cite{devlin2019bert} are released under the Apache License 2.0. GPT-2~\cite{radford2019language} is released under the Modified MIT License. MedGemma~\cite{sellergren2025medgemma} is released under the Health AI Developer Foundations Terms of Use. All pretrained models are accessed via the Hugging Face Transformers library~\cite{wolf-etal-2020-transformers}, which is also released under the Apache License 2.0.


\end{document}